%% file: main.tex
\documentclass[10pt,twocolumn,letterpaper]{article}
\usepackage[T1]{fontenc}
\usepackage[pagenumbers]{cvpr}
\input{preamble}
\definecolor{cvprblue}{rgb}{0.21,0.49,0.74}
\usepackage[pagebackref,breaklinks,colorlinks,allcolors=cvprblue]{hyperref}
\def\confName{CVPR}
\def\confYear{2027}

\title{VGR-Searcher: Enhancing Visually Grounded Reasoning \\ for Multimodal Search Agents}
\author{%
  Zixian Guo$^{1,2,*}$, Feng Zhu$^{1,*}$, Rongzhi Zhang$^{3}$, Jiarui Zhao$^{3}$, Lingchuan Liu$^{3,\dagger}$,\\
  Fan Yang$^{3}$, Xunliang Cai$^{3}$, Lei Zhang$^{2}$, Wangmeng Zuo$^{1,\dagger}$\\[4pt]
  {\normalsize $^{1}$Harbin Institute of Technology \quad $^{2}$The Hong Kong Polytechnic University \quad $^{3}$Meituan}%
}
\AddToHookNext{shipout/foreground}{%
  \put(\LenToUnit{1in},\LenToUnit{-0.68in}){\includegraphics[width=3cm]{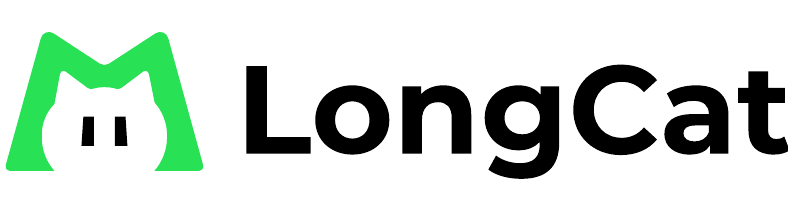}}%
}
\begin{document}
\maketitle
{\renewcommand{\thefootnote}{\ensuremath{*}}%
 \footnotetext{Equal contribution.}}
{\renewcommand{\thefootnote}{\ensuremath{\dagger}}%
 \footnotetext{Corresponding authors.}}
\input{sec/0_abstract}
\input{sec/1_intro}
\input{sec/2_related_work}
\input{sec/3_method}
\input{sec/4_experiments}

\section{Conclusion}
We presented VGR-Searcher, a multimodal search agent trained with a scalable pipeline for constructing questions that require sustained visual reasoning. Starting from a multimodal Wikipedia graph, our pipeline derives discriminative visual clues by comparing semantically related images and composes them with textual reasoning steps. Complementary branches are iteratively refined so that each admits multiple candidates and their combination uniquely identifies the answer within the inspected pool. These designs require close visual inspection and evidence integration across searches. Supervised fine-tuning on teacher demonstrations followed by reinforcement learning yields a new state-of-the-art across ten evaluated multimodal search benchmarks. On visually intensive tasks like BrowseComp-V$^3$, VGR-Searcher exceeds the strongest prior 8--9B specialized search agent by a large margin of 14.7 percentage points. The model also generalize well to text-only search tasks. We will release the full pipeline for question construction, expert trajectory collection, and model training to help others reproduce our results.

\newpage
\section*{LLM Usage Statement}
The authors used Codex to assist with manuscript preparation, including language editing, grammar correction, and improving clarity. The authors formulated the research questions, developed the methodology, designed and organized the experiments, conducted the evaluation and analysis, and drew the scientific conclusions. They reviewed, revised, and verified all AI-assisted outputs and take full responsibility for the submitted content.

{\small
\bibliographystyle{ieeenat_fullname}
\bibliography{main}
}
\input{sec/X_suppl}

\end{document}

%% file: preamble.tex
\usepackage{array}
\usepackage{colortbl}
\usepackage{stfloats}

\newcommand{\suppl}{\textit{Appendix}}

\usepackage{tikz}
\usepackage{tcolorbox}
\definecolor{promptnavy}{HTML}{20364C}
\newtcolorbox{systempromptbox}[1][System prompt]{
  colback=white,colframe=promptnavy,colbacktitle=promptnavy,
  coltitle=white,title={#1},fonttitle=\sffamily\bfseries,
  fontupper=\small\ttfamily,boxrule=0.4pt,arc=2pt,
  left=7pt,right=7pt,top=7pt,bottom=7pt,
  before skip=8pt,after skip=8pt
}

%% file: sec/0_abstract.tex
\begin{abstract}
With recent advances in training data synthesis, multimodal deep search agents can integrate interleaved textual and visual evidence to guide subsequent searches and answer complex questions.
Yet many questions in existing training datasets require little further visual reasoning once the agent identifies the entity depicted in the image.
To encourage visual evidence gathering and multimodal reasoning, we introduce a novel data synthesis pipeline for questions that require agents to continuously resolve visual clues across successive searches.
Specifically, we compare semantically related images to construct visual clues that distinguish the target image from plausible alternatives. To follow the clue, the agent must conduct close visual inspection and identify further search and reasoning directions.
To increase the complexity of the reasoning paths, we further construct questions with complementary reasoning chains so that each chain alone leaves several plausible answers, while their combination uniquely determines the answer. Together, these mechanisms encourage close visual inspection and evidence integration across multiple searches.
We use these questions to train a 9B multimodal search agent, \textbf{VGR-Searcher}, through supervised fine-tuning on teacher demonstrations followed by reinforcement learning.
VGR-Searcher outperforms prior specialized search agents on all ten visual search benchmarks, with a substantial lead of 14.7 percentage points on BrowseComp-V$^3$.
We will release the full stack of data construction and model training to support reproducible research on multimodal agentic search.
\end{abstract}

%% file: sec/1_intro.tex
\vspace{-3pt}
\section{Introduction}
\label{sec:introduction}

Advances in tool use and multi-turn reasoning have enabled language-model agents to pursue complex queries through successive web searches~\cite{li2025websailor,li2025webthinker,wu2025webdancer}. Multimodal web search is more challenging because the evidence needed to answer a question is distributed across web pages and images.
Agents search for relevant images and examine the visual evidence by closely inspecting and comparing fine-grained details. After identifying a visual clue, they integrate it with earlier findings from other sources, drawing on this interleaved textual and visual evidence to determine the next search query or resolve the question.
Human-authored benchmarks such as MMSearch-Plus~\cite{tao2026mmsearchplus}, MM-BrowseComp~\cite{li2025mmbrowsecomp}, and BrowseComp-V$^3$~\cite{zhang2026browsecompv3} provide challenging testbeds for multimodal agentic search, with questions that can require dozens of tool interactions to trace visual clues, inspect relevant images, and integrate visual and textual evidence over long reasoning processes.

Human authors can design benchmark questions around several visual clues and verify complex evidence dependencies, but doing so at the scale of a training set is costly. 
This has motivated recent work on automated data construction~\cite{chu2026redsearcher,huang2026vision}. However, the integration of intensive visual reasoning in search tasks is still limited.
For example, a common approach replaces an entity name in a textual reasoning chain with an image~\cite{geng2025webwatcher}. Once the agent recognizes the depicted entity, the remaining steps degrade into text-only search.
More recent works~\cite{zhang2026deepvoyager,he2026neos} introduce dependencies on newly retrieved images in subsequent searches, yet such visual dependencies do not necessarily require non-trivial visual reasoning. Individual steps can still be solved through straightforward visual recognition without closely inspecting the retrieved images. Furthermore, none of these existing approaches provide enough details for a reproducible framework.

These limitations highlight three key goals for scalable data construction: (1) \textbf{multi-step visual reasoning}, where each step requires non-trivial visual interpretation beyond simple entity recognition; (2) \textbf{complementary visual evidence}, where clues obtained from separate searches contribute distinct information that must be integrated to derive the final answer; and (3) \textbf{reproducible and scalable construction}, enabled by an open-source, systematic pipeline for consistent question generation and verification.
To meet these goals, we introduce a scalable data synthesis pipeline. We draw on textual relations and illustration images from Wikipedia to build a multimodal knowledge graph. Along a path through this graph, a textual clue can guide image retrieval, and a visual observation can help determine which entity to investigate next, providing a foundation for constructing reasoning paths. We use these paths to construct individual steps with discriminative visual clues and to compose complementary reasoning chains.

Specifically, we construct discriminative visual clues by comparing images with related semantic content. For each proposed visual connection in the graph, we retrieve alternatives to its supporting image and inspect them alongside the target. The comparison identifies observable differences in attributes, actions, or spatial relations that distinguish the target from plausible alternatives. We express these differences as clues in the question. To resolve a clue during question solving, an agent must examine the retrieved images, compare candidates against the specified visual details, and determine which entity satisfies the description. This construction provides supervision for visual discrimination among relevant candidates along the reasoning path.

To make the reasoning structure more challenging, we introduce branching structures composed of complementary reasoning chains. Each branch combines textual relations with visual conditions to narrow the candidate answers while retaining multiple plausible candidates. Identifying the answer requires the agent to jointly reason over textual and visual evidence gathered through separate searches and reconcile the constraints imposed by both branches. During construction, we verify that the branches jointly identify a unique answer within the inspected candidate pool and that removing either branch's visual condition restores ambiguity. These questions require close visual inspection and evidence from both searches.

To collect expert demonstrations for these questions, we build an agent harness that lets a teacher model search the web, read webpages, and retrieve and inspect images in detail.
During collection, the harness checks for errors such as malformed tool calls immediately after each turn, and resamples the entire turn when errors are detected.
We use these demonstrations for supervised fine-tuning (SFT) of a 9B vision-language model, then train it further with reinforcement learning (RL).
The resulting multimodal search agent, VGR-Searcher, achieves 48.2\% accuracy on MM-BrowseComp and 52.0\% on BrowseComp-V$^3$, exceeding the best results previously reported for 8--9B specialized search agents according to Table~\ref{tab:main_results}. The lead on BrowseComp-V$^3$ is 14.7 percentage points. Our contributions are threefold.

\begin{itemize}
    \item We introduce a scalable data synthesis pipeline for constructing multimodal search questions designed to match the difficulty of human-authored questions. By requiring visual evidence to distinguish among candidates and combining complementary reasoning chains, the resulting questions encourage close visual inspection and evidence integration throughout the search process.
    \item VGR-Searcher outperforms prior specialized agents of comparable size across ten multimodal search benchmarks. Although trained without text-only search data, it also generalizes well on all three tested text-based search benchmarks.
    \item We document the detailed configuration for building visual reasoning, the collection of expert trajectories, and the training method. The full stack will be released as a reproducible recipe for multimodal search agents.
\end{itemize}

%% file: sec/2_related_work.tex
\section{Related Work}
\label{sec:related}

\vspace{-1pt}
\noindent \textbf{Multimodal deep search agents.}
Training multimodal search agents involves learning both to acquire evidence through tools and to use it in subsequent decisions. Existing methods address these requirements through complementary forms of supervision. SFT transfers search behavior from expert demonstrations, making question construction and trajectory quality central to agent development: WebWatcher uses synthetic multimodal trajectories for cold-start training~\cite{geng2025webwatcher}, while DeepVoyager-VL learns visual acquisition and long-horizon search through supervised training alone~\cite{zhang2026deepvoyager}. RL further optimizes decisions through interaction feedback. MMSearch-R1 combines outcome rewards with a search penalty to learn when and how to retrieve evidence~\cite{wu2025mmsearchr1}; WebWatcher follows SFT with RL, and Neos uses reference-guided credit assignment and joint adaptation of the policy and evidence interface~\cite{geng2025webwatcher,he2026neos}. Across these training strategies, the questions determine which evidence-seeking behaviors receive supervision, motivating our focus on constructing tasks that require sustained visual reasoning.

\vspace{-1pt}
\noindent \textbf{Constructing visual reasoning tasks.}
A central challenge in task construction is making visual evidence necessary beyond initial entity recognition. WebWatcher replaces entity mentions in textual reasoning chains with image references~\cite{geng2025webwatcher}; once the depicted entities are recognized, such questions may reduce to textual search. Subsequent approaches increase either the complexity of the search structure or the role of visual evidence within it. REDSearcher controls graph topology and evidence dispersion and introduces visual cues into search tasks~\cite{chu2026redsearcher}. Vision-DeepResearch connects visual evidence acquisition with long-horizon textual reasoning~\cite{huang2026vision}, while Visual-Seeker injects questions about retrieved visual details~\cite{zhang2026visualseeker}. DeepVoyager-VL introduces intermediate visual bindings in event-graph reasoning programs, and Neos uses visual checkpoints, distractors, and rewriting to reduce textual shortcuts~\cite{zhang2026deepvoyager,he2026neos}. These developments shift task design toward dependencies that require visual inspection during search, rather than merely visual inputs.

However, reproducing these constructions requires engineering details not fully specified in the papers, and limited access to complete pipelines makes it difficult to regenerate or scale the training questions. Publicly released resources for learning visual evidence integration include training data from REDSearcher and Vision-DeepResearch~\cite{chu2026redsearcher,huang2026vision}; our data ablation shows that training on questions from these releases yields lower accuracy than training on our constructed questions (Table~\ref{tab:data_source}). Our pipeline supports reproducible question construction at scale by deriving discriminative visual clues and verifying complementary branches on a multimodal Wikipedia graph.

\noindent \textbf{Benchmarks for multimodal deep search.}
Early benchmarks evaluated multimodal retrieval and factual question answering grounded in input images~\cite{jiang2024mmsearch,wu2025mmsearchr1,fu2025livevqa,cheng2025simplevqa}. Recent multimodal deep search benchmarks extend this evaluation to the use of visual evidence across retrieval steps. BrowseComp-VL combines visual identification with BrowseComp-style multi-hop research~\cite{geng2025webwatcher}, while MMSearch-Plus asks agents to trace fine-grained visual cues to their sources through successive searches~\cite{tao2026mmsearchplus}. Expert-authored questions in MM-BrowseComp and VisBrowse-Bench require agents to retrieve and integrate visual evidence from the web~\cite{li2025mmbrowsecomp,zhang2026visbrowse}. BrowseComp-V$^3$ structures visually dependent questions around verifiable subgoals~\cite{zhang2026browsecompv3}. VDR-Bench combines region-based visual retrieval, multi-hop expansion, and expert checks to limit textual leakage and trivial image matching~\cite{zeng2026vdrbench}. Experts design and verify these questions to ensure that the clues jointly identify the answer and that shortcuts cannot bypass the intended reasoning. The cost of verifying each question makes manual construction impractical at training scale, motivating our automated construction of questions with explicit evidence dependencies.

%% file: sec/3_method.tex
\begin{figure*}[!t]
\centering
\includegraphics[width=.985\textwidth]{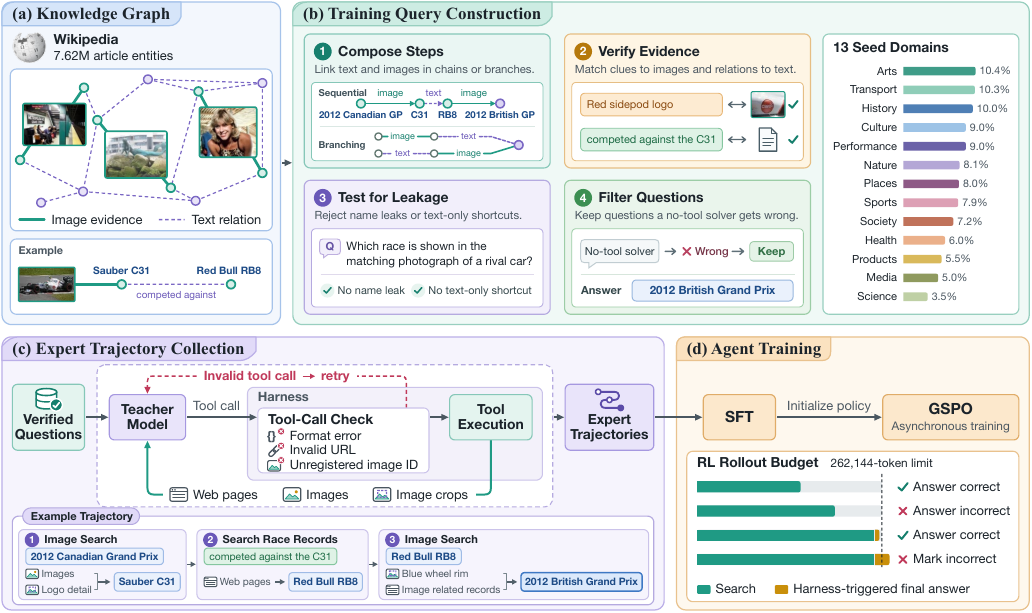}
\vspace{-4pt}
\caption{Training pipeline for VGR-Searcher. (a, b) We compose visual and textual steps from a multimodal Wikipedia graph, verify their dependencies, and filter questions for entity leakage and tool-free solvability. The chart shows the domain distribution of the training questions. (c, d) We train the agent with SFT on teacher demonstrations followed by RL with GSPO.}
\vspace{-1pt}
\label{fig:framework}
\end{figure*}

\section{Method}
\label{sec:method}
We train VGR-Searcher to use visual evidence to guide a search across multiple steps. We describe how the agent interacts with web pages and images, then how we construct training questions around evidence drawn from Wikipedia. A teacher model solves the verified questions to provide demonstrations for supervised fine-tuning (SFT), after which the agent is finetuned through reinforcement learning (RL). Figure~\ref{fig:framework} summarizes our method.

\begin{figure*}[t]
\centering
\includegraphics[width=\textwidth]{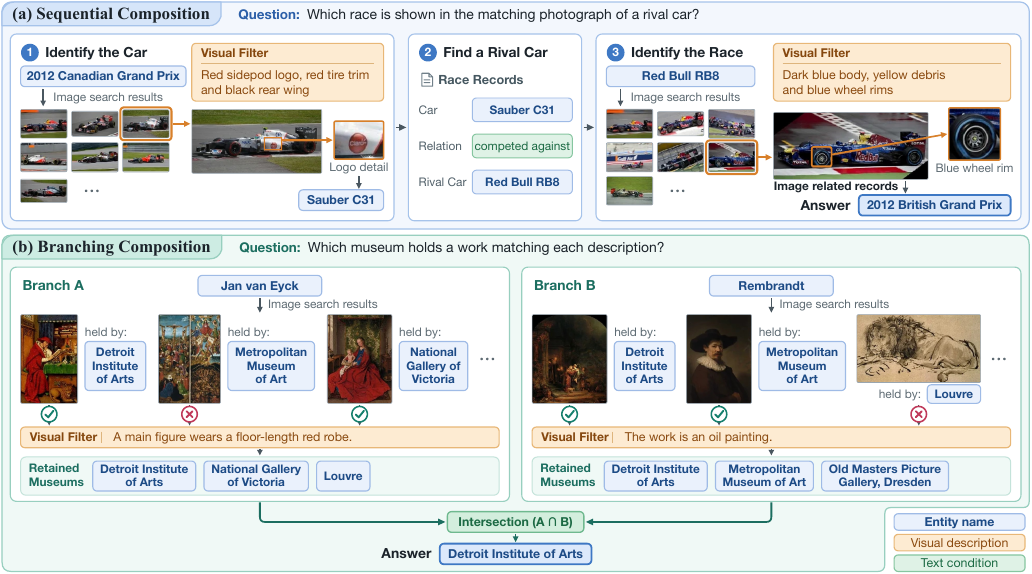}
\caption{Examples of question construction. (a) Sequential composition links two visual steps through a textual relation. A red sidepod logo helps identify the Sauber C31. Its competitive relationship with the Red Bull RB8 leads to a second image comparison that identifies the 2012 British Grand Prix. (b) In branching composition, each artwork clue retains three candidate museums. Only the Detroit Institute of Arts appears in both sets within the inspected image pools.}
\label{fig:query_construction}
\end{figure*}

\subsection{Problem Formulation and Agent Interface}
\label{sec:harness}

Given a question, the agent uses search tools to gather the evidence it needs to answer.
Let $h_0$ denote the initial context containing the question. At turn $t$, the search policy generates an action $a_t \sim \pi_\theta(\cdot \mid h_t)$. A final answer ends the interaction. For a tool-use action, the harness parses the requested tools and arguments, executes the calls, and appends the action and its observations to the context:
\begin{equation}
    o_t = \mathrm{Execute}\bigl( \mathrm{Parse}(a_t)\bigr), 
    h_{t+1} = h_t \oplus (a_t, o_t),
\end{equation}
where $\oplus$ denotes message concatenation. Interleaved text and image observations guide the agent's next search.

In addition to the search and webpage-reading tools used by text-based agents, our harness provides \texttt{ImageSearch} and \texttt{VisualSearch} to retrieve images from textual queries and visual references. \texttt{ImageRead} loads selected image pixels into the model's context, while \texttt{ImageZoomIn} exposes a cropped region for detailed inspection. These tools allow the agent to search for visual evidence, inspect it alongside textual observations, and use the findings to guide subsequent searches. See \suppl~\ref{app:harness} for the full details of the agent harness.

\subsection{Scalable Question Construction}
\label{sec:question_construction}

In a complex search question, an entity identified in one source can guide the solver's search for evidence in another. To find such connections at scale, we build a multimodal knowledge graph that links Wikipedia text and images through shared entities. We use the graph to construct visual and textual reasoning steps, compose them into sequential or branching questions, and verify that each question preserves the intended dependencies between steps.

\subsubsection{Knowledge Graph Construction}
\label{sec:representation}

\noindent \textbf{Textual relations.}
Following prior work on generating search questions from knowledge graphs~\cite{li2025websailor,du2026openseeker,zhao2026lohosearch}, we build a textual graph from the full English Wikipedia dump. Each article represents an entity, and links in its body form directed edges to the articles they reference. This graph contains approximately 7.62 million entities and 265 million directed edges. To guide sampling, we record entity categories from Wikidata~\cite{wikidata2014denny} when available and use in-degree as a proxy for entity popularity.

\noindent \textbf{Connections through visual evidence.}
We scan illustrations in Wikipedia articles and use hyperlinks in their captions to identify the entities associated with each image. The image provides visual evidence for the relation between these entities and is stored with the corresponding edge and its source article. The resulting visual graph shares entity nodes with the textual graph and contains 1.89 million entities and 5.13 million directed edges.

\subsubsection{Reasoning Step Construction}
\label{sec:atomic}

A graph edge records a connection between known entities. To turn that connection into a reasoning step, we leave the target entity unnamed and write a clue that points to it. We construct visual steps that require the solver to inspect candidate images and textual steps that require the solver to trace a relation described in a source passage. Each clue and its target form one reasoning step. We combine these steps into a question.

\noindent \textbf{Visual reasoning steps.}
A visual step uses observable details to narrow the set of candidate entities. An overly specific description can disclose an entity through the wording alone. We therefore show the question-generation model the image supporting a graph edge alongside semantically related alternatives (Fig.~\ref{fig:query_construction}a). The model selects a few details, such as a pose or the relative positions of objects, that distinguish the target from some alternatives while leaving its identity implicit. At search time, the agent inspects candidate images for these details and uses their sources to identify the entities supported by the matching images. Subsequent steps or branches resolve residual ambiguity.

\noindent \textbf{Textual reasoning steps.}
For an edge in the textual graph, we extract a relation description from the source entity's Wikipedia page. We phrase the relation from source to target, leaving the target unnamed so that the agent must identify it from textual evidence. We retain the supporting passage to verify the relation and its direction.

\subsubsection{Reasoning Step Composition}
\label{sec:composition}

To construct questions that require the solver to combine evidence from multiple searches, we organize visual and textual reasoning steps into sequential or branching structures.

\noindent \textbf{Sequential composition.}
We compose visual and textual steps into a chain in which the entity resolved at one step provides the starting point for the next. Given a desired number of steps, we assemble a question skeleton that records the starting point given to the solver, the entity transitions and their associated clues, and the expected answer type.
A writing model links the step descriptions into a natural-language question that asks for the final entity and refers to intermediate entities through their clues.

\noindent \textbf{Branching composition.}
\label{sec:branches}
We construct branching questions by composing two reasoning chains that provide complementary clues to the same target. One way to generate such questions~\cite{zhang2026visualseeker} is to give a question-generation model a branching structure. In our initial attempts with this approach, however, the model sometimes omitted one branch or included both even when a single branch sufficed to identify the answer. To make both branches necessary, we introduce an iterative procedure that compares the candidate answers supported by each branch and revises the clues when needed. We retain a branch pair only if each branch leaves several plausible answers within its candidate pool and the target is the only answer common to both.

For a target $a^\star$, we select two distinct public search anchors. For each branch $b$, let $U_b$ contain the candidate entities supported by sources in its inspected image pool, and let $C_b\subseteq U_b$ contain those supported by an image matching the branch's visual description. We compare canonical entity identifiers so that aliases count once. Each branch must leave at least two plausible entities, while their intersection identifies the target:
\begin{equation}
    |C_1|\geq2,\quad |C_2|\geq2,\quad C_1\cap C_2=\{a^\star\}.
    \label{eq:joint}
\end{equation}
Even when both branches are necessary, the nonvisual information in one branch might already suffice to narrow the intersection to the target. We therefore remove each visual condition in turn, keeping its anchor and identity relation fixed. In each case, at least one alternative must reappear:
\begin{equation}
    |U_1\cap C_2|\geq2,\qquad |C_1\cap U_2|\geq2.
    \label{eq:ablation}
\end{equation}
If a check fails, the model revises the descriptions and we recompute the candidate sets. We allow one joint revision and up to four narrowing attempts. Each narrowing step retains the target and at least one alternative within the previous set. Entities remain fixed unless their evidence fails verification. We discard any structure that still fails.

For an accepted structure, the writing model phrases each branch separately using only its public conditions and the expected answer type. We combine the two descriptions into one question asking which entity satisfies both, then attach the reference answer. The shared entity can also serve as the starting point for further steps.

In both sequential and branching questions, we require visual evidence in the final step, drawing on prior findings on its use in multimodal search~\cite{zhang2026visualseeker,huang2026vision}.

\begin{table*}[t]
\centering
\small
\caption{Trajectory length and tool use across question sources. For each dataset, we report averages over trajectories generated by DeepSeek-V4.1-Flash-1017. Each query in a batched \texttt{Search} call counts as one call.}
\vspace{-2pt}
\label{tab:training_tool_use}
\setlength{\tabcolsep}{3pt}
\resizebox{\textwidth}{!}{\begin{tabular}{@{}l*{7}{c}|ccc@{}}
\toprule
Question source & \multicolumn{7}{c|}{Mean calls of each tool per trajectory} & \multicolumn{3}{c}{Mean length per trajectory}\\
\cmidrule(lr){2-8}\cmidrule(lr){9-11}
Training set or benchmark & Search & ScholarSearch & \textbf{ImageSearch} & \textbf{VisualSearch} & \textbf{ImageRead} & \textbf{ImageZoomIn} & LinkReader & Turns & Calls & Tokens\\
\midrule
REDSearcher-MM & 17.87 & 0.17 & 0.37 & 0.45 & 0.68 & 0.41 & 5.36 & 9.3 & 25.3 & 36.3~k\\
Vision-DeepResearch & 9.29 & 0.55 & 1.27 & 1.24 & 2.37 & 1.68 & 4.27 & 9.6 & 20.7 & 27.9~k\\
VGR-Searcher SFT training set & 33.84 & 0.07 & 5.79 & 0.06 & 6.31 & 0.43 & 10.29 & 23.4 & 56.8 & 59.9~k\\
\midrule
MMSearch-Plus & 30.30 & 0.19 & 6.94 & 2.30 & 8.20 & 9.84 & 12.77 & 31.7 & 70.6 & 91.3~k\\
BrowseComp-V$^3$ & 35.56 & 0.10 & 8.64 & 2.06 & 10.88 & 9.67 & 18.63 & 40.1 & 85.5 & 120.3~k\\
\bottomrule
\end{tabular}}
\end{table*}

\subsubsection{Question Verification}
\label{sec:realization}

Putting the clues into a single question can inadvertently reveal an intermediate entity or obscure a dependency between steps. We therefore verify the final wording against the intended reasoning structure and its supporting evidence. The question-generation model checks each visual clue against the images and each textual relation against its source passage. We reject descriptions that name a hidden entity or reveal its identity through a paraphrase. For each visual step, the model tests whether the public anchor and relation wording alone identify the next entity. If so, it cites the relevant wording and explains the inference, and we discard the question. The model also checks that the final wording preserves the required clues, relation directions, and dependencies in the skeleton. Finally, a solver receives only the complete question and answers without tools. An independent grader compares the answer with the reference. We retain only questions with a completed, incorrect blind answer.

\subsection{Expert Trajectory Collection}
\label{sec:training}

We collect questions evenly across 13 domains independently of the evaluation benchmarks and construct expert demonstrations by having the teacher model solve the verified questions with the agent harness. Figure~\ref{fig:framework} shows the collected SFT data by domain. A collected trajectory with a correct final answer can mask tool-related errors in the actions that SFT teaches the student to imitate. In an initial collection with the original test-time harness without tool-call checks, we checked 3,609 correctly answered trajectories for malformed arguments, empty search queries, references to unregistered images, and inaccessible URLs unsupported by the preceding context. Filtering all flagged trajectories would discard up to 40.1\% of otherwise correct demonstrations. Retaining them exposes the student to failed tool calls as supervision.

We therefore introduce a tool validation module in the harness during trajectory collection to intercept obvious tool call errors mentioned above. The tool check intervenes at the turn level. If the tool call in a turn fails validation, the teacher resamples this turn from the same prior context, up to three times after the initial attempt. Stochastic decoding in thinking mode allows the teacher to produce different actions without a corrective message in its context. We preserve the original context across retries instead of prompting the model to revise the tool calling to minimize interference with the teacher's reasoning. Benchmark evaluation runs without collection-time tool checks or regeneration.

With these checks and retries, we observed little change in the teacher's final-answer accuracy during expert trajectory generation. But the proportion of correctly answered trajectories with tool issues has been reduced to 9.8\%. We retain complete demonstrations with semantically correct answers and at most 128k tokens for SFT, check image accessibility and tool structure, and standardize the final-answer format. Table~\ref{tab:training_tool_use} shows that the solver searched for and inspected images more often on our questions than on questions from the two public training sets. \suppl~\ref{app:collection_guard} details the checks and selection criteria.

\input{sec/main_results}

\subsection{Agent Training}
\label{sec:agent_training}

\noindent \textbf{Supervised initialization.}
We perform SFT on the retained demonstrations, using the teacher's reasoning, tool calls, and final answer as prediction targets. The question and tool observations provide conditioning context and are masked from the loss. This trains the student to acquire and interpret evidence through complete search interactions. Frameworks, hyperparameters, and final-answer formatting for SFT and RL are detailed in \suppl~\ref{app:training_details}.

\noindent \textbf{Reinforcement learning.}
We select RL questions from a pool disjoint from the questions used for SFT. Before training, we sample four rollouts per candidate question from the SFT policy and retain the question only if one to three rollouts yield a correct answer. When all four answers are correct or all four are incorrect, the query can be too easy or too hard to supply useful contrast for the group-normalized advantage. We then optimize the policy on the retained questions using Group Sequence Policy Optimization (GSPO)~\cite{zheng2025gspo}. We use terminal rewards for answer correctness, without intermediate process supervision.

\noindent \textbf{Rollout context budget.}
During RL rollouts, the SFT model often exhausted its 128k-token context while searching, leaving no room for a final answer. We therefore increase the rollout budget to 256k tokens and reserve the last 10k for the answer. When another search turn would enter this reserve, the harness inserts the \texttt{<final\_summary>} prefix so the agent can answer using the evidence already collected. We score the rollout as incorrect if the agent calls another tool or runs out of context before giving an answer.

%% file: sec/main_results.tex
\newcommand{\scoresd}[2]{#1\textsubscript{\textpm#2}}
\begin{table*}[!t]
\centering
\caption{Multimodal search benchmark scores (\%). Bold and underlined scores denote the best and second-best results within each comparison group. The VGR-Searcher rows report mean accuracy over three runs, with the sample standard deviation across runs in subscript. For published search agents, $^{*}$ marks scores from our evaluations, and $^{\dagger}$ marks models not publicly available.}
\vspace{-2pt}
\label{tab:main_results}
\setlength{\tabcolsep}{3pt}\scriptsize
\resizebox{\textwidth}{!}{\begin{tabular}{@{}l*{5}{c}|*{5}{c}|c@{}}
\toprule
Model & MMSearch & BC-VL & SimpleVQA & LiveVQA & FVQA & VDR & MMSearch+ & MM-BC & VisBrowse & BC-V$^3$ & Overall\\

\midrule\multicolumn{12}{@{}l}{\emph{Direct Answer}}\\
GPT-6.1 Sol~\cite{openai2026gpt61sol} & \underline{71.0} & \underline{51.1} & \underline{68.0} & \underline{72.3} & \underline{61.3} & \textbf{19.6} & \textbf{28.6} & \underline{18.3} & \textbf{41.4} & \underline{20.7} & \underline{45.2}\\
DeepSeek-V4.1-Flash-1017~\cite{deepseek2026flash1017} & 57.0 & 45.9 & 64.3 & 66.3 & 50.7 & 10.4 & 9.3 & 15.6 & 28.4 & 14.0 & 36.2\\
Gemini 4 Carbon~\cite{google2026gemini4carbon} & 55.3 & 34.6 & 63.0 & 49.0 & 50.3 & 10.0 & 10.3 & 12.9 & 23.1 & 13.0 & 32.2\\
Gemini 4 Argon~\cite{google2026gemini4argon} & 55.0 & 35.8 & 65.0 & 51.3 & 52.0 & 11.8 & 13.2 & 14.3 & 24.3 & 14.7 & 33.7\\
Kimi K3.1 Pro~\cite{moonshot2026kimik31pro} & \textbf{74.7} & \textbf{52.4} & \textbf{78.7} & \textbf{75.7} & \textbf{72.0} & \underline{19.2} & \underline{26.7} & \textbf{21.9} & \underline{39.1} & \textbf{26.3} & \textbf{48.7}\\
Qwen3.5-9B~\cite{qwen2026nativeagents} & 21.7 & 28.6 & 58.3 & 53.3 & 28.3 & 5.0 & 1.6 & 8.0 & 10.7 & 8.3 & 22.4\\
\midrule\multicolumn{12}{@{}l}{\emph{Multimodal Search Agent Harness}}\\
GPT-6.1 Sol & 86.7 & 69.2 & 85.3 & 89.0 & 83.0 & 43.4 & 66.2 & \underline{61.2} & 72.2 & \underline{64.7} & 72.1\\
DeepSeek-V4.1-Flash-1017 & \textbf{92.7} & \textbf{76.4} & \underline{87.7} & \underline{90.3} & \underline{87.0} & \underline{47.6} & \underline{69.8} & 59.4 & \underline{72.8} & 63.3 & \underline{74.7}\\
Gemini 4 Carbon & 84.0 & 59.4 & 82.0 & 84.0 & 82.7 & 34.4 & 42.1 & 23.2 & 36.1 & 25.7 & 55.4\\
Gemini 4 Argon & 84.7 & 61.9 & 83.3 & 84.0 & 83.0 & 35.2 & 44.7 & 22.8 & 34.3 & 26.3 & 56.0\\
Kimi K3.1 Pro & \underline{91.7} & \underline{75.2} & \textbf{89.7} & \textbf{95.0} & \textbf{88.7} & \textbf{50.2} & \textbf{72.3} & \textbf{71.4} & \textbf{81.1} & \textbf{70.0} & \textbf{78.5}\\
Qwen3.5-9B & 78.0 & 54.6 & 74.2 & 72.3 & 73.7 & 26.9 & 27.2 & 21.7 & 31.8 & 22.2 & 48.3\\
\midrule\multicolumn{12}{@{}l}{\emph{Multimodal Deep Search Agents}}\\
MM-DeepResearch-8B~\cite{yao2026mmdeepresearch} & 67.8 & 37.9 & 65.9 & 65.0 & 69.2 & ${}^{*}$14.2 & ${}^{*}$13.2 & ${}^{*}$9.6 & ${}^{*}$18.3 & ${}^{*}$9.3 & 37.0\\
OpenSearch-VL-8B~\cite{chen2026opensearch} & 64.5 & 37.6 & 71.6 & 59.6 & 71.5 & 20.8 & ${}^{*}$17.8 & ${}^{*}$8.5 & ${}^{*}$18.6 & ${}^{*}$11.3 & 38.2\\
SearchEyes-9B$^{\dagger}$~\cite{jiao2026searcheyes} & 69.2 & 42.3 & 75.4 & 66.1 & 74.2 & 28.3 & -- & -- & -- & -- & --\\
DeepVoyager-VL-8B$^{\dagger}$~\cite{zhang2026deepvoyager} & 72.7 & 58.4 & 76.3 & 82.7 & 82.7 & 35.0 & 37.1 & 24.0 & 47.3 & 32.3 & 54.8\\
OneSearch-VL-8B~\cite{li2026onesearch} & 66.5 & 39.0 & 72.4 & 61.2 & 73.4 & 23.6 & ${}^{*}$17.2 & ${}^{*}$4.0 & ${}^{*}$9.5 & ${}^{*}$8.2 & 37.5\\
Neos-9B$^{\dagger}$~\cite{he2026neos} & 81.3 & 63.2 & \underline{81.3} & 88.7 & 83.3 & 31.4 & 47.6 & 39.5 & 53.3 & 37.0 & 60.7\\
MMSearch-OPD-9B$^{\dagger}$~\cite{zhang2026mmsearchopd} & \underline{86.3} & 67.7 & -- & \underline{89.0} & -- & -- & 47.0 & 42.0 & 52.1 & 37.3 & --\\
ModalSearcher-8B$^{\dagger}$~\cite{yin2026modalsearcher} & 70.3 & 48.1 & -- & 70.3 & 69.9 & 26.2 & -- & -- & 33.4 & 21.3 & --\\
\rowcolor[gray]{0.93}
VGR-Searcher-9B (SFT) & \scoresd{85.7}{0.67} & \scoresd{\underline{70.3}}{1.59} & \scoresd{80.8}{0.39} & \scoresd{88.2}{0.84} & \scoresd{\underline{83.8}}{1.95} & \scoresd{\underline{42.3}}{1.90} & \scoresd{\underline{52.0}}{1.45} & \scoresd{\underline{47.5}}{2.06} & \scoresd{\underline{61.9}}{1.81} & \scoresd{\underline{50.0}}{1.46} & \scoresd{\underline{66.2}}{0.13}\\
\rowcolor[gray]{0.93}
VGR-Searcher-9B (RL) & \scoresd{\textbf{88.3}}{1.34} & \scoresd{\textbf{71.6}}{0.63} & \scoresd{\textbf{83.4}}{1.54} & \scoresd{\textbf{90.4}}{0.39} & \scoresd{\textbf{87.0}}{0.88} & \scoresd{\textbf{45.4}}{0.69} & \scoresd{\textbf{55.5}}{3.58} & \scoresd{\textbf{48.2}}{1.18} & \scoresd{\textbf{67.3}}{1.81} & \scoresd{\textbf{52.0}}{4.04} & \scoresd{\textbf{68.9}}{0.77}\\
\bottomrule
\end{tabular}}
\end{table*}

%% file: sec/4_experiments.tex
\section{Experiments}
\label{sec:experiments}

\subsection{Experimental Setup}

\noindent \textbf{Benchmarks.}
We evaluate multimodal search on ten benchmarks, grouped as in Table~\ref{tab:main_results}. The first group covers earlier multimodal search benchmarks: MMSearch~\cite{jiang2024mmsearch}, BrowseComp-VL (BC-VL)~\cite{geng2025webwatcher}, SimpleVQA~\cite{cheng2025simplevqa}, LiveVQA~\cite{fu2025livevqa}, and FVQA~\cite{wu2025mmsearchr1}. The second emphasizes more intensive visual evidence integration: VDR-Bench (VDR)~\cite{zeng2026vdrbench}, MMSearch-Plus (MMSearch+)~\cite{tao2026mmsearchplus}, MM-BrowseComp (MM-BC)~\cite{li2025mmbrowsecomp}, VisBrowse-Bench (VisBrowse)~\cite{zhang2026visbrowse}, and BrowseComp-V$^3$ (BC-V$^3$)~\cite{zhang2026browsecompv3}. We also evaluate text-based search benchmarks on BrowseComp~\cite{wei2025browsecomp}, BrowseComp-ZH~\cite{zhou2025browsecompzh}, and DeepSearchQA~\cite{gupta2026deepsearchqa} for more generalized evaluation.

\noindent \textbf{Baselines and evaluation.}
We compare the SFT and RL checkpoints of VGR-Searcher with three groups of baselines (Table~\ref{tab:main_results}).
In \emph{Direct Answer}, general-purpose models answer questions without external tools.
In \emph{Agentic Workflow}, these models search the web and inspect retrieved text and images using the same harness as our SFT and RL models.
For prior \emph{Multimodal Deep Search Agents}, we use reported scores for closed-source agents and evaluate open-source agents on unreported benchmarks using their own harness that their model had been trained on.
We use the standard judging pipeline from prior work~\cite{huang2026vision,he2026neos,zhang2026mmsearchopd} to measure final-answer accuracy, with Qwen3.6-35B-A3B as the judge. For our SFT and RL models, we run inference independently three times on every question and report the mean and sample standard deviation of accuracy across the three runs. Overall accuracy is the unweighted mean across the ten benchmarks, computed separately for each run. Evaluation and training details appear in \suppl~\ref{app:harness} and \suppl~\ref{app:training_details}, respectively.

\subsection{Main Results}

\noindent \textbf{Multimodal search.}
Under the agentic workflow, the Qwen3.5-9B base model~\cite{qwen2026nativeagents} averages 48.3\% across the ten benchmarks. SFT on our expert demonstrations raises this average to 66.2\%, and RL further improves it to 68.9\% (Table~\ref{tab:main_results}). VGR-Searcher presents a new state-of-the-art with a large margin, especially on benchmarks requiring complex reasoning over visual evidence.
The RL model consistently improves on SFT across all ten benchmarks.
On BrowseComp-V$^3$, it reaches 52.0\%, compared to 37.3\% for MMSearch-OPD-9B, the strongest prior specialized agent.
These gains support our data and training choices.

\noindent \textbf{Transfer to text-based search.}
For text-based search, we use the same system prompt and seven tools as in multimodal evaluation. Despite receiving no text-only search training data, VGR-Searcher improves over the Qwen3.5-9B base model on all three benchmarks (Table~\ref{tab:text_search}). The largest gain is on BrowseComp-ZH, where accuracy rises from 32.9\% to 59.0\%. Across all three text-based benchmarks, the agent made no calls to any of the four visual tools, indicating the model generalizes well to different tasks with adaptive tool selection.

\begin{table}[t]
\centering
\footnotesize
\caption{Text-based search accuracy (\%). Qwen3.5-9B and VGR-Searcher-9B use the same agent harness.}
\vspace{-2pt}
\label{tab:text_search}
\setlength{\tabcolsep}{2pt}
\begin{tabular}{@{}lccc@{}}
\toprule
Model & \shortstack{Browse\\Comp} & \shortstack{Browse\\Comp-ZH} & \shortstack{Deep\\SearchQA}\\
\midrule
Qwen3.5-9B~\cite{qwen2026nativeagents} & 22.2 & 32.9 & 41.0\\
Marco-DR-8B~\cite{zhu2026marco} & 31.4 & \underline{47.1} & 29.2\\
REDSearcher-MM-30B~\cite{chu2026redsearcher} & 31.2 & 44.5 & --\\
DeepSearch-World-9B~\cite{geng2026deepsearchworld} & 31.2 & 36.4 & --\\
Neos-9B~\cite{he2026neos} & \underline{36.4} & 45.3 & \textbf{54.3}\\
\midrule
VGR-Searcher-9B & \textbf{44.8} & \textbf{59.0} & \underline{54.2}\\
\bottomrule
\end{tabular}
\end{table}

\subsection{Ablation Studies}
\label{sec:question_sources}

\noindent \textbf{Search demands of our training questions.}
We ask whether our training questions elicit the sustained search and visual inspection seen on challenging benchmarks. To examine this, we compare trajectories generated by DeepSeek-V4.1-Flash-1017~\cite{deepseek2026flash1017} on our questions, two public training sets, and two evaluation benchmarks (Table~\ref{tab:training_tool_use}). The solver averaged 23.4 turns on our questions, up from 9.3 on REDSearcher-MM and 9.6 on Vision-DeepResearch, but still below the benchmark averages of 31.7 and 40.1 turns.
On our questions, the solver averaged 33.84 \texttt{Search} queries and 10.29 \texttt{LinkReader} calls per trajectory, compared with 9.29--17.87 queries and 4.27--5.36 calls on the public training sets. It also searched for and inspected images more often on our questions, averaging 5.79 \texttt{ImageSearch} and 6.31 \texttt{ImageRead} calls per trajectory, compared with 0.37--1.27 and 0.68--2.37, respectively. 

\begin{table}[t]
\centering
\caption{SFT training-data ablation (accuracy, \%).}
\vspace{-2pt}
\label{tab:data_source}
\footnotesize
\setlength{\tabcolsep}{2.5pt}
\begin{tabular}{@{}lccc@{}}
\toprule
SFT training data & MMSearch+ & MM-BC & BC-V$^3$\\
\midrule
VDR (845) & 48.4 & 34.4 & 37.3\\
REDSearcher-MM (1,984) & 45.5 & 37.1 & 41.0\\
VDR (845) + REDSearcher-MM (1,139) & 47.4 & 35.7 & 40.7\\
\midrule
Sequential + Branching (Overall 1,984) & \textbf{52.0} & \textbf{47.5} & \textbf{50.0}\\
\bottomrule
\end{tabular}
\end{table}

\noindent \textbf{SFT training data.}
To assess the effect of the training questions on SFT model performance, we finetune Qwen3.5-9B on demonstrations collected with the same teacher model and agent harness (Table~\ref{tab:data_source}). Each configuration uses 1,984 demonstrations except for VDR~\cite{huang2026vision} alone, since VDR only released 1,000 questions, and the teacher answered 845 correctly. We therefore compare against two public-data baselines of equal size: REDSearcher-MM alone and the 845 VDR demonstrations supplemented with 1,139 from REDSearcher-MM. Training on our mixture of sequentially composed and branching questions yields 52.0\% accuracy on MMSearch-Plus, 47.5\% on MM-BrowseComp, and 50.0\% on BrowseComp-V$^3$. These scores exceed the stronger of the two size-matched baselines on each benchmark by 4.6, 10.4, and 9.0 percentage points, respectively.
The demonstration trajectories collected on our constructed questions have more practice combining search and visual tools within a trajectory.



\begin{table}[t]
\centering
\caption{RL versus further SFT for VGR-Searcher on the same training questions. Scores are mean accuracies (\%) over three runs, with sample standard deviations in subscript.}
\vspace{-2pt}
\label{tab:rl_value}
\small
\setlength{\tabcolsep}{3pt}
\begin{tabular}{@{}lccc@{}}
\toprule
Benchmark & SFT & SFT (+ RL data) & SFT + RL\\
\midrule
MMSearch-Plus & \scoresd{52.0}{1.45} & \scoresd{\underline{53.5}}{0.32} & \scoresd{\textbf{55.5}}{3.58}\\
MM-BrowseComp & \scoresd{\underline{47.5}}{2.06} & \scoresd{45.9}{2.87} & \scoresd{\textbf{48.2}}{1.18}\\
BrowseComp-V$^3$ & \scoresd{\underline{50.0}}{1.46} & \scoresd{48.3}{2.73} & \scoresd{\textbf{52.0}}{4.04}\\
\bottomrule
\end{tabular}
\end{table}

\noindent \textbf{Effectiveness of reinforcement learning.}
During multimodal search, each tool response changes the evidence available for the next decision. An agent may need to revise a query, inspect another image, or decide that it has enough evidence to answer. SFT teaches these decisions along expert trajectories. RL lets the agent act on the observations produced by its own choices and uses the final outcome to update the policy. 
To compare RL with further SFT on the same questions, we collect teacher demonstrations for the RL questions and add them to the SFT training set (Table~\ref{tab:rl_value}). The results reveal that additional supervised data does not provide consistent benefit. It improves mean accuracy on MMSearch-Plus but lowers it on the other two benchmarks.  SFT followed by RL yields the highest mean accuracy on all three benchmarks, exceeding further SFT by 2.0 to 3.7 percentage points. This comparison supports the use of outcome-guided interaction to improve performance on held-out search tasks.
The dynamics of reward, interaction counts, and tool use during agentic RL are detailed in \suppl~\ref{app:rl_dynamics}.

%% file: sec/X_suppl.tex
\clearpage
\appendix
\setcounter{table}{0}
\renewcommand{\thetable}{S\arabic{table}}
\input{sec/appendix}

%% file: sec/appendix.tex
\noindent\textbf{Guide to the appendix.}
Section~\ref{app:harness} describes the agent interface, evaluation protocol, and teacher-trajectory collection.
Section~\ref{app:training_details} gives the SFT and RL settings and compares SFT results by question structure.
Section~\ref{app:rl_dynamics} tracks reward and tool use during RL. Section~\ref{app:zero_rl} evaluates RL initialized directly from Qwen3.5-9B.
Section~\ref{app:question_construction} illustrates how we construct and verify sequential and branching questions.
Section~\ref{app:search_behavior} compares search behavior across benchmarks, and Section~\ref{app:qualitative_search} illustrates visual tool use in individual trajectories.
Section~\ref{app:code_data} lists the accompanying code and example questions.

\section{Agent Interface and Implementation Details}
\label{app:harness}

This section details the seven-tool agent interface, system prompt, context-budget control, and teacher-trajectory collection. We add tool-call validation and turn regeneration during teacher-trajectory collection.

\noindent \textbf{Evaluation protocol.}
For benchmark evaluation, Qwen3.6-35B-A3B compares each final answer with the reference answer and generates its judgment at temperature~0. BrowseComp-VL accuracy pools L1 and L2 by question count, and VDR uses the 500-question mini split.

\noindent \textbf{Knowledge graph construction.}
We construct the directed hyperlink graph from the March 2025 full English Wikipedia dump~\cite{wikimediaenwikidumps}. To obtain entity types for sampling, we query the Wikidata SPARQL endpoint for the available \emph{instance of} (P31) values of each entity~\cite{wikidata2014denny}.

\subsection{Tool Set and Visual Observations}
\label{app:tools}

The agent uses the seven tools in Table~\ref{tab:appendix_tools}. Tool schemas are supplied through the model's function-calling interface, separately from the system prompt. Calls use JSON arguments, and each tool response is associated with its originating call through a \texttt{tool\_call\_id}. Images are registered under identifiers such as \texttt{<image: 0>}. We use Serper~\cite{serperapi} for web, scholarly, image, and reverse image search. An identifier or URL provides a reference to an image; \texttt{ImageRead} and \texttt{ImageZoomIn} supply image pixels as multimodal observations for direct perception.

\begin{table*}[t]
\centering
\small
\caption{The seven-tool interface. An asterisk marks an optional argument.}
\label{tab:appendix_tools}
\begin{tabular}{@{}>{\raggedright\arraybackslash}p{0.16\textwidth}>{\raggedright\arraybackslash}p{0.29\textwidth}>{\raggedright\arraybackslash}p{0.49\textwidth}@{}}
\toprule
Tool & Arguments & Function and returned evidence\\
\midrule
\texttt{Search} & \texttt{query}: list of strings & Searches the web for textual queries. Accepts up to five queries per request and also supports a single string. Returns search results for locating relevant sources.\\[3pt]
\texttt{ScholarSearch} & \texttt{query}: string & Retrieves scholarly papers, reports, and related sources for a focused research query.\\[3pt]
\texttt{ImageSearch} & \texttt{query}: string & Retrieves images from a textual query, including image references and source-page URLs.\\[3pt]
\texttt{VisualSearch} & \texttt{reference\_id}\newline\texttt{region\_of\_interest}$^{*}$\newline\texttt{thumbnail\_size}$^{*}$ & Performs reverse image search with a registered image or a selected region. Returns visual matches and their sources.\\[3pt]
\texttt{ImageRead} & \texttt{reference\_id} & Downloads the image and loads its pixels into context for inspection of objects, text, and other details.\\[3pt]
\texttt{ImageZoomIn} & \texttt{reference\_id}, \texttt{region} & Crops and loads a specified region into context for detailed inspection. The crop is supplied as an image observation.\\[3pt]
\texttt{LinkReader} & \texttt{url}, \texttt{description} & Uses WebBrowse to fetch the page or document at the supplied URL and extracts evidence relevant to the stated information need. The schema takes one URL; the collection harness additionally supports URL lists.\\
\bottomrule
\end{tabular}
\end{table*}

\begin{table*}[!t]
\centering\small
\caption{Tool issues among teacher trajectories with correct final answers. Percentages use the correct trajectories in each collection run as the denominator. A trajectory may appear in multiple categories and counts once in the final row.}
\label{tab:teacher_tool_issues}
\begin{tabular}{@{}>{\raggedright\arraybackslash}p{0.54\textwidth}cc@{}}
\toprule
Tool issue & Evaluation harness & With validation and retries\\
\midrule
Malformed tool call (e.g., invalid JSON, missing required arguments, or an empty search query) & 8.89\% & 4.80\%\\[3pt]
URL whose domain was absent from the preceding context and whose tool response explicitly reported an access failure & 12.99\% & 4.10\%\\[3pt]
Invalid image reference (nonexistent image ID or out-of-range index) & 23.73\% & 4.80\%\\
\midrule
\textbf{At least one of the above} & \textbf{40.09\%} & \textbf{9.76\%}\\
\bottomrule
\end{tabular}
\end{table*}

\noindent \textbf{Image coordinates and identifiers.}
Both \texttt{region} and \texttt{region\_of\_interest} are serialized strings containing four normalized coordinates, \texttt{[min\_x, min\_y, max\_x, max\_y]}. The optional search region defaults to the full image. \texttt{thumbnail\_size} accepts \texttt{tiny}, \texttt{small}, \texttt{medium}, or \texttt{large}, corresponding to long-side sizes of 112, 224, 448, or 896 pixels; the default is \texttt{medium}. The agent can inspect a full image or a region and use what it sees to guide a text or image search.

\noindent \textbf{Image references across turns.}
For each question, the harness maintains an image registry (\texttt{image\_map}) shared by the tools. It assigns each image an identifier such as \texttt{<image: 1>} and associates it with its source URL or encoded image data. The agent sees the identifier alongside an input image or in an image search result and can pass it as \texttt{reference\_id} in a later call. For example, \texttt{ImageRead} resolves the identifier and returns the image pixels, while \texttt{ImageZoomIn} returns pixels from the requested region. The harness coordinates identifier assignment across parallel calls and saves the registry with the conversation so that references still resolve to the same images when a trajectory resumes.

\subsection{Complete System Prompt}
\label{app:system_prompt}

We use the following system prompt for both multimodal and text-based search. The harness sends it in the system message and supplies the seven tool definitions in the request's \texttt{tools} field. For our Qwen model served by vLLM, the model's chat template renders the prompt, tool schemas, and tool-calling instructions as the system portion of the input, followed by the conversation history.

\begin{systempromptbox}
\raggedright
You are an agent designed to accomplish tasks.

Your goal is to provide a comprehensive, objective, and well-researched answer to the user's query. You have access to multiple tools. Please read each tool's description and parameters carefully. There is no limit on how many times tools can be used. Please reason and respond in the same language as the user's question, unless the user explicitly requests otherwise. Put your final answer between \textless{}answer\textgreater{} and \textless{}/answer\textgreater{}.
\end{systempromptbox}

The harness accepts answers enclosed in \texttt{<answer>} tags and, for VGR-Searcher, \texttt{<final\_summary>} tags. The latter support the context-budget mechanism described below.

\begin{table*}[t]
\centering\small
\caption{Supervised fine-tuning settings for VGR-Searcher-9B.}
\label{tab:sft_settings}
\begin{tabular}{@{}ll@{}}
\toprule
Setting & Value\\
\midrule
Initialization and framework & Qwen3.5-9B, MS-SWIFT (Megatron backend)\\
Trainable parameters & Full language backbone, with frozen vision encoder and aligner\\
Epochs / global batch size / microbatch size & 4 / 32 / 1\\
Peak / minimum learning rate & $10^{-5}$ / $10^{-6}$\\
Learning-rate schedule & 5\% warmup, cosine decay\\
Adam $(\beta_1,\beta_2,\epsilon)$ & $(0.9,0.95,10^{-8})$\\
Weight decay / gradient clipping norm & 0.1 / 1.0\\
Maximum sequence length & 128k tokens\\
Supervised tokens & Assistant reasoning, tool calls, and final answer\\
Precision and attention & BF16, Flash Attention\\
Batch representation & Padding-free, with sequence packing disabled\\
Tensor / pipeline / context parallel sizes & 4 / 2 / 1, with sequence parallelism enabled\\
Activation recomputation & Selective (core attention and MLP)\\
\bottomrule
\end{tabular}
\end{table*}

\begin{table*}[!t]
\centering\small
\caption{Reinforcement learning settings for VGR-Searcher-9B.}
\label{tab:rl_settings}
\begin{tabular}{@{}ll@{}}
\toprule
Setting & Value\\
\midrule
\multicolumn{2}{@{}l}{\textit{Policy optimization}}\\
Initial policy / RL questions & SFT checkpoint 322 / 4,196\\
Framework / algorithm & verl (Megatron backend) / GSPO\\
Trainable parameters & Language backbone; vision encoder and visual projection frozen\\
Optimizer / learning-rate schedule & Adam / constant $10^{-6}$, no warmup\\
Adam $(\beta_1,\beta_2,\epsilon)$ & $(0.9,0.98,10^{-8})$\\
Weight decay / gradient clipping norm & 0.1 / 1.0\\
Policy minibatch / microbatch per device & 256 trajectories (32 groups of 8) / 1 trajectory\\
Policy clipping bounds & $1-4\times10^{-4}$ and $1+4\times10^{-4}$\\
Loss aggregation & Sequence mean of token means\\
KL / entropy regularization & Both disabled\\
Updates per rollout batch & 1\\
Advantage normalization & Within-question mean and standard deviation\\
Reward / training judge & Binary answer correctness / Qwen2.5-32B-Instruct\\
Policy loss mask & Generated tokens only, excluding the injected answer prefix\\
RL updates for the reported model & 77\\
\midrule
\multicolumn{2}{@{}l}{\textit{Rollout generation}}\\
Prompts per batch / trajectories per prompt & 32 / 8\\
Context budget / prefill trigger margin & 262,144 / 10,240 tokens\\
Maximum generation per call & 16,384 tokens\\
Turn reminder threshold & 200 assistant turns (the agent may continue)\\
Rollout timeout & 7,200 seconds\\
Sampling temperature / top-$p$ / top-$k$ & 1.0 / 1.0 / disabled\\
Precision & BF16\\
Weight synchronization & Every policy update\\
Staleness parameter & 1.5\\
Rollout importance weighting & Token-level, capped at 2.0\\
Random seed & 42\\
\bottomrule
\end{tabular}
\end{table*}

\subsection{Final-Answer Generation and Context Budgets}
\label{app:final_summary}

During RL rollouts and benchmark evaluation, a long search can exhaust the context window before the agent produces a complete answer. To leave room for an answer based on the evidence gathered so far, the harness switches from search to final-answer generation when fewer than 10,240 tokens remain in the 262,144-token window. It prefills the next assistant response with the following text.
\begin{quote}\begin{minipage}{\linewidth}\small\ttfamily\raggedright
\textless think\textgreater{}\par
The context budget is nearly exhausted. I will answer the original question now using the evidence already collected.\par
\textless/think\textgreater{}\par\medskip
\textless final\_summary\textgreater{}
\end{minipage}\end{quote}
The model continues directly with its final answer, and the adapter restores the prefix when parsing the response. This response ends the search. The mechanism is disabled by default during teacher collection. During RL, injected prefix tokens are excluded from the policy loss. Trajectories that exhaust the context budget before producing a final answer receive reward 0, as detailed in Section~\ref{app:training_details}.

\subsection{Reliable Teacher-Trajectory Collection}
\label{app:collection_guard}

To improve tool reliability in SFT demonstrations, we validate tool calls while collecting teacher trajectories. Table~\ref{tab:teacher_tool_issues} compares two collections on training questions using the same teacher model. Among trajectories with correct final answers, the share containing a tool error fell from 40.09\% with the standard evaluation harness to 9.10\% with validation and retries.

\noindent\textbf{Tool validation rules.} The harness flags the following calls:
\begin{itemize}
    \item \textbf{Invalid tools or arguments.} The tool is unavailable, its arguments cannot be parsed as a JSON object, or a search request contains no nonempty query.
    \item \textbf{Invalid image references.} The reference does not resolve to a registered image, as with an unknown image ID or an out-of-range index.
    \item \textbf{Unverified URLs.} The URL points to a domain absent from the preceding interaction history and an attempted read explicitly reports an access failure.
    \item \textbf{Explicit database queries.} The URL encodes a SQL or SPARQL query or points to the Wikidata query endpoint.
\end{itemize}

\label{app:trajectory_filtering}
If none of the tool calls in a turn passes validation, the teacher regenerates that turn from the same history without a corrective prompt. Stochastic decoding in thinking mode allows the teacher to produce a valid call on a subsequent attempt. The harness allows up to three retries after the initial attempt and executes the final attempt if all retries fail. We retain complete trajectories with correct final answers and at most 128k tokens for SFT. In each retained demonstration, we wrap the teacher's final answer in \texttt{<final\_summary>} and \texttt{</final\_summary>} tags, preserving its wording. This gives the student examples of completing an answer from the same opening tag used during RL and evaluation.

\begin{table*}[t]
\centering
\small
\caption{Mean turns, tool calls, and final-call tokens in correct teacher trajectories by question source and structure.}
\label{tab:search_complexity}
\begin{tabular}{@{}lccc@{}}
\toprule
Question source / structure & Turns & Tool calls & Tokens\\
\midrule
REDSearcher-MM & 11.2 & 17.9 & 43.6~k\\
Vision-DeepResearch & 11.2 & 19.0 & 33.8~k\\
Sequential (two steps) & 23.3 & 35.2 & 55.2~k\\
Sequential (three steps) & 37.3 & 56.9 & 94.5~k\\
Branching & 29.2 & 46.8 & 70.3~k\\
\bottomrule
\end{tabular}
\end{table*}

\begin{table*}[t]
\centering
\small
\caption{SFT accuracy (\%) by training data source and question structure. The best score in each column is bold and the second best is underlined.}
\label{tab:suppl_sft_data}
\begin{tabular}{@{}lcccc@{}}
\toprule
SFT training data & Demonstrations & MMSearch-Plus & MM-BrowseComp & BrowseComp-V$^3$\\
\midrule
VDR & 845 & 48.4 & 34.4 & 37.3\\
REDSearcher-MM & 1,984 & 45.5 & 37.1 & 41.0\\
VDR (845) + REDSearcher-MM (1,139) & 1,984 & 47.4 & 35.7 & 40.7\\
\midrule
Sequential only & 1,984 & \underline{51.9} & 46.0 & \underline{49.3}\\
Branching only & 1,085 & 51.0 & \underline{47.3} & 46.0\\
Sequential + branching & 1,984 & \textbf{52.0} & \textbf{47.5} & \textbf{50.0}\\
\bottomrule
\end{tabular}
\end{table*}

\begin{figure*}[t]
\centering
\includegraphics[width=\textwidth]{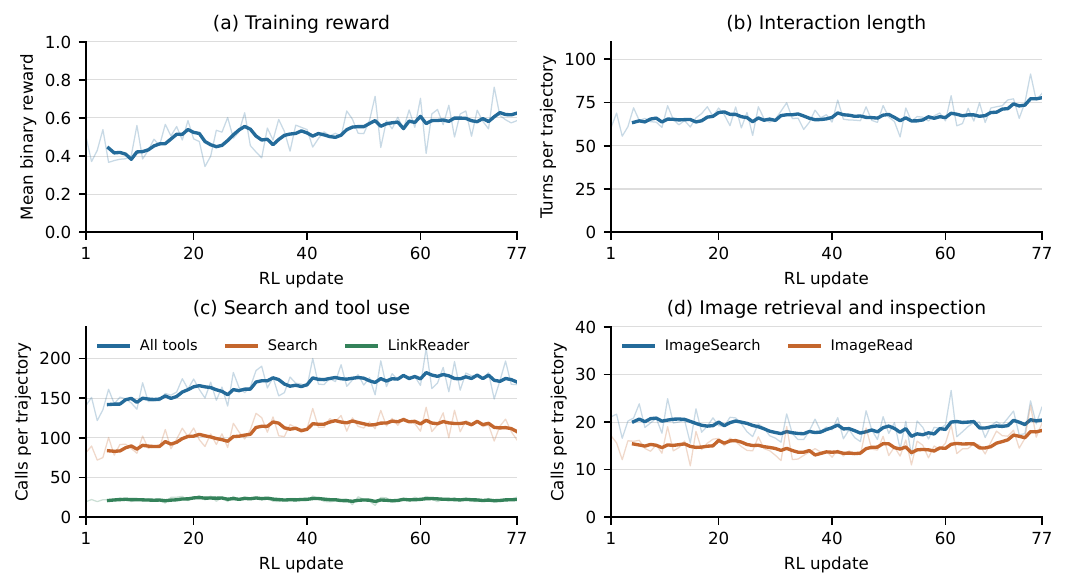}
\caption{Reward and search behavior during RL training through checkpoint 77. Thin lines show per-step means; thick lines show trailing five-step means. Reward uses valid training trajectories, while interaction statistics use complete tool records. Each of the 77 RL updates uses a batch of 256 trajectories. Batched \texttt{Search} queries count separately.}
\label{fig:rl_dynamics}
\end{figure*}

\section{Training Details}
\label{app:training_details}

\noindent \textbf{Supervised fine-tuning.}
We fine-tune the Qwen3.5-9B language backbone on the teacher demonstrations using MS-SWIFT~\cite{zhao2025swift} (Table~\ref{tab:sft_settings}). The loss covers the teacher's reasoning, tool calls, and final answer. The question and tool observations serve as context and are masked from the loss.

\noindent \textbf{Reinforcement learning.}
Starting from the SFT checkpoint, we train on 4,196 questions using GSPO in verl~\cite{sheng2025hybridflow} (Table~\ref{tab:rl_settings}). For each question, we sample eight trajectories and normalize their terminal rewards within the group to compute advantages. Each batch of 32 groups yields one policy update. We evaluate the policy after 77 updates.

\noindent \textbf{Reward and loss.}
Given the question, reference answer, and final response, Qwen2.5-32B-Instruct classifies the response as correct, incorrect, or not attempted. A correct response receives reward 1. Incorrect responses, unattempted questions, and trajectories that exhaust the context budget receive reward 0. We exclude trajectories interrupted by infrastructure or grader failures from both group statistics and the loss. The policy loss covers only tokens generated by the agent, excluding tool observations and the prefix supplied by the harness.

\noindent \textbf{Asynchronous rollouts.}
We synchronize rollout weights after each policy update. Ongoing searches can span an update, so we use token-level importance weights to account for differences between the rollout and training policies. Long searches use the final-answer prefill described in Section~\ref{app:final_summary}.

\subsection{Trajectory Statistics and SFT Results by Question Structure}
\label{app:trajectory_complexity}
Table~\ref{tab:search_complexity} summarizes all correctly answered trajectories in the original teacher collections before subsampling and the 128k-token training length filter. Turns include the final answer, and parallel tool calls count separately. Final-call tokens are the input and output tokens reported by the teacher API for its last response. The token means cover 3,105 of 3,106 three-step trajectories and 1,212 of 1,213 branching trajectories, and every trajectory from the other sources.

\noindent \textbf{Effect of question structure on fine-tuning.}
Table~\ref{tab:suppl_sft_data} compares Qwen3.5-9B fine-tuned on demonstrations collected with the same teacher and agent harness. The sequential and mixed configurations use 1,984 demonstrations each, and the branching configuration uses 1,085. The sequential configuration achieves higher accuracy than the branching configuration on MMSearch-Plus and BrowseComp-V$^3$. Branching questions yield higher accuracy on MM-BrowseComp (47.3\% versus 46.0\%). The mixture achieves the highest observed accuracy on all three benchmarks, exceeding the better single-structure result by 0.1, 0.2, and 0.7 percentage points, respectively.

\section{Reward and Tool Use During RL Training}
\label{app:rl_dynamics}
Figure~\ref{fig:rl_dynamics} tracks reward and search behavior over 77 RL updates. Reward is averaged over trajectories eligible for policy optimization, including context-budget failures assigned reward 0. Turn and tool statistics use trajectories with complete interaction records. Each query in a batched \texttt{Search} request counts separately; other tool requests count once, including unsuccessful requests.

\noindent \textbf{Reward and search behavior.}
Comparing the first ten RL updates (1--10) with the final ten (68--77), mean reward increases from 0.432 to 0.603. Over the same windows, the mean number of assistant turns rises from 63.5 to 75.9 and the mean number of tool calls rises from 143.4 to 174.7. Image retrieval and inspection remain frequent: \texttt{ImageSearch} averages 20.1 and 20.4 calls in the two windows, while \texttt{ImageRead} increases from 15.0 to 17.7.

\noindent \textbf{Context-budget behavior.}
The context limit remains fixed at 262,144 tokens throughout the run. Among logged training episodes, 34.1\% of episodes activate the final-answer prefill, while 1.0\% terminate with a context-budget failure. The mean observed context length is approximately 175.5k tokens.

\section{Reinforcement Learning from the Base Model}
\label{app:zero_rl}

To assess the value of supervised initialization, we also apply RL directly to the Qwen3.5-9B base model. We call this setting Zero-RL. With the same search harness, Zero-RL improves overall accuracy from 48.3\% to 58.1\% (Table~\ref{tab:zero_rl}). Even with this gain, Zero-RL trails SFT alone on nine of the ten benchmarks. The sole exception is SimpleVQA, where Zero-RL scores 82.3\% and SFT scores 80.8\%. Initializing the policy with SFT on teacher demonstrations and then applying RL yields the highest accuracy on all ten benchmarks, reaching 68.9\% overall.

\begin{table*}[t]
\centering
\caption{Accuracy (\%) on ten multimodal search benchmarks. Subscripts give sample standard deviations across three runs for our models. Overall weights the ten benchmarks equally within each run. Bold and underlined values mark the highest and second-highest scores among trained search agents. As in Table~\ref{tab:main_results}, $^{*}$ marks our evaluations of existing agents and $^{\dagger}$ denotes models not publicly available. Within MMSearch-Plus, Zero-RL scores \scoresd{55.7}{4.03}\% on the easy subset and \scoresd{35.2}{3.07}\% on the difficult subset.}
\label{tab:zero_rl}
\setlength{\tabcolsep}{3pt}\scriptsize
\resizebox{\textwidth}{!}{\begin{tabular}{@{}l*{5}{c}|*{5}{c}|c@{}}
\toprule
Model & MMSearch & BC-VL & SimpleVQA & LiveVQA & FVQA & VDR & MMSearch+ & MM-BC & VisBrowse & BC-V$^3$ & Overall\\
\midrule
Qwen3.5-9B (harness) & 78.0 & 54.6 & 74.2 & 72.3 & 73.7 & 26.9 & 27.2 & 21.7 & 31.8 & 22.2 & 48.3\\
\midrule
MM-DeepResearch-8B~\cite{yao2026mmdeepresearch} & 67.8 & 37.9 & 65.9 & 65.0 & 69.2 & ${}^{*}$14.2 & ${}^{*}$13.2 & ${}^{*}$9.6 & ${}^{*}$18.3 & ${}^{*}$9.3 & 37.0\\
OpenSearch-VL-8B~\cite{chen2026opensearch} & 64.5 & 37.6 & 71.6 & 59.6 & 71.5 & 20.8 & ${}^{*}$17.8 & ${}^{*}$8.5 & ${}^{*}$18.6 & ${}^{*}$11.3 & 38.2\\
SearchEyes-9B$^{\dagger}$~\cite{jiao2026searcheyes} & 69.2 & 42.3 & 75.4 & 66.1 & 74.2 & 28.3 & -- & -- & -- & -- & --\\
DeepVoyager-VL-8B$^{\dagger}$~\cite{zhang2026deepvoyager} & 72.7 & 58.4 & 76.3 & 82.7 & 82.7 & 35.0 & 37.1 & 24.0 & 47.3 & 32.3 & 54.8\\
OneSearch-VL-8B~\cite{li2026onesearch} & 66.5 & 39.0 & 72.4 & 61.2 & 73.4 & 23.6 & ${}^{*}$17.2 & ${}^{*}$4.0 & ${}^{*}$9.5 & ${}^{*}$8.2 & 37.5\\
Neos-9B$^{\dagger}$~\cite{he2026neos} & 81.3 & 63.2 & 81.3 & 88.7 & 83.3 & 31.4 & 47.6 & 39.5 & 53.3 & 37.0 & 60.7\\
MMSearch-OPD-9B$^{\dagger}$~\cite{zhang2026mmsearchopd} & \underline{86.3} & 67.7 & -- & \underline{89.0} & -- & -- & 47.0 & 42.0 & 52.1 & 37.3 & --\\
ModalSearcher-8B$^{\dagger}$~\cite{yin2026modalsearcher} & 70.3 & 48.1 & -- & 70.3 & 69.9 & 26.2 & -- & -- & 33.4 & 21.3 & --\\
\midrule
\rowcolor[gray]{0.93}
VGR-Searcher-9B (SFT) & \scoresd{85.7}{0.67} & \scoresd{\underline{70.3}}{1.59} & \scoresd{80.8}{0.39} & \scoresd{88.2}{0.84} & \scoresd{\underline{83.8}}{1.95} & \scoresd{\underline{42.3}}{1.90} & \scoresd{\underline{52.0}}{1.45} & \scoresd{\underline{47.5}}{2.06} & \scoresd{\underline{61.9}}{1.81} & \scoresd{\underline{50.0}}{1.46} & \scoresd{\underline{66.2}}{0.13}\\
\rowcolor[gray]{0.93}
VGR-Searcher-9B (SFT + RL) & \scoresd{\textbf{88.3}}{1.34} & \scoresd{\textbf{71.6}}{0.63} & \scoresd{\textbf{83.4}}{1.54} & \scoresd{\textbf{90.4}}{0.39} & \scoresd{\textbf{87.0}}{0.88} & \scoresd{\textbf{45.4}}{0.69} & \scoresd{\textbf{55.5}}{3.58} & \scoresd{\textbf{48.2}}{1.18} & \scoresd{\textbf{67.3}}{1.81} & \scoresd{\textbf{52.0}}{4.04} & \scoresd{\textbf{68.9}}{0.77}\\
\rowcolor[gray]{0.93}
VGR-Searcher-9B (Zero-RL) & \scoresd{85.2}{0.77} & \scoresd{62.1}{0.77} & \scoresd{\underline{82.3}}{1.45} & \scoresd{81.0}{0.67} & \scoresd{81.8}{0.69} & \scoresd{34.1}{1.94} & \scoresd{41.4}{1.34} & \scoresd{33.3}{1.13} & \scoresd{46.6}{2.92} & \scoresd{33.2}{0.77} & \scoresd{58.1}{0.28}\\
\bottomrule
\end{tabular}}
\end{table*}

\section{Question Construction and Verification}
\label{app:question_construction}
Figure~\ref{fig:query_construction} illustrates how we construct a visual reasoning step and how we combine clues from two branches to identify a common answer. Here we follow both examples from their source images to the visual conditions and complete questions presented to the agent.

\subsection{Constructing Visual Conditions from Candidate Images}

We start with an image that connects two entities in the Wikipedia graph (Figure~\ref{fig:framework}). Using captions and linked entities, we select semantically related images associated with alternative entities. We give the question generator the target image and a contact sheet of these alternatives. The following prompt asks it to identify details that a solver could recognize among the retrieved images.

\begin{systempromptbox}[Question construction prompt]
\raggedright\textbf{Sequential visual clue (excerpt)}\par
Describe only the first image. Use the second image to choose a concise but sufficiently informative combination of visual details that clearly distinguishes the target from the supplied distractors, not to produce an exhaustive caption. Prioritize clarity and reliable discrimination over brevity.
\end{systempromptbox}

In the racing example, the first comparison yields a white and black car with \emph{a red circular logo on its white side panel, red trim on all its tires, and a solid black rear wing}. These details help the solver locate the photograph whose source identifies the Sauber C31. We repeat the comparison for a rival car. Here the distinguishing features are \emph{fine yellow granular debris} on a dark blue body and \emph{blue-accented tire rims}. The source of the matching photograph identifies the race. Thus the first visual step returns a car model, and the last returns an event.

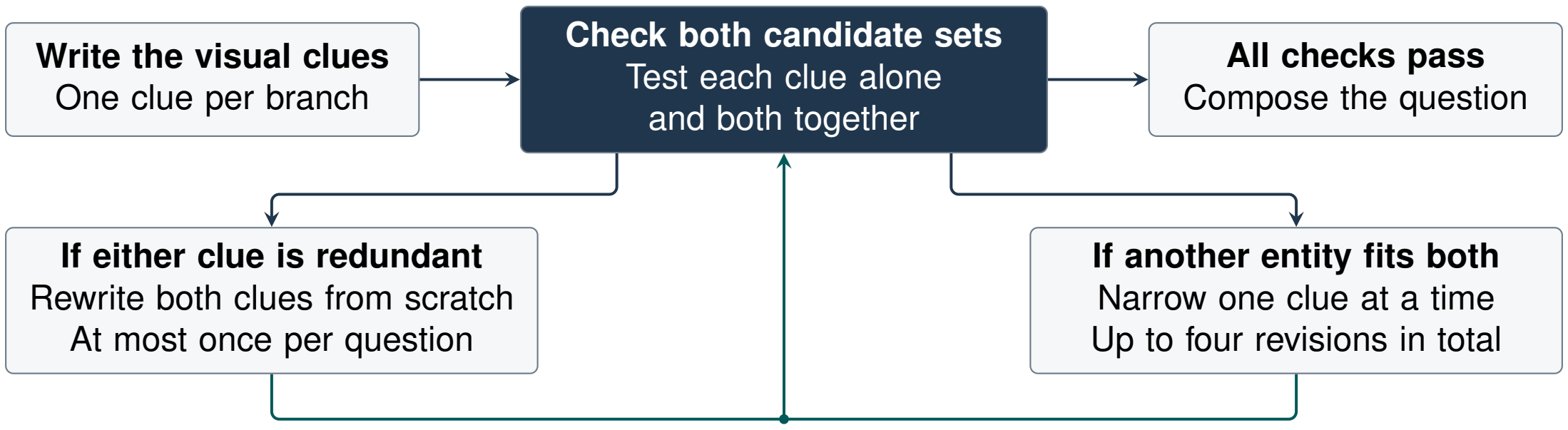
\begin{figure*}[t]
\centering\small
\begin{tikzpicture}[
    x=1cm,y=1cm,font=\small\sffamily,
    box/.style={draw=promptnavy!65,rounded corners=2pt,align=center,
                inner sep=4pt,minimum height=1.05cm,fill=promptnavy!4},
    edge/.style={draw=promptnavy,line width=.75pt,rounded corners=2pt},
    arrow/.style={edge,->,>=stealth},
    back/.style={edge,draw=teal!70!black}]
\node[box,text width=3.55cm] (start) at (-5.3,0)
    {\textbf{Write the visual clues}\\One clue per branch};
\node[box,text width=4.6cm,fill=promptnavy,text=white] (check) at (0,0)
    {\textbf{Check both candidate sets}\\Test each clue alone\\and both together};
\node[box,text width=3.55cm] (pass) at (5.3,0)
    {\textbf{All checks pass}\\Compose the question};
\node[box,text width=4.65cm] (both) at (-4.75,-2.05)
    {\textbf{If either clue is redundant}\\Rewrite both clues from scratch\\At most once per question};
\node[box,text width=4.65cm] (one) at (4.75,-2.05)
    {\textbf{If another entity fits both}\\Narrow one clue at a time\\Up to four revisions in total};
\draw[arrow] (start.east) -- (check.west);
\draw[arrow] (check.east) -- (pass.west);
\draw[arrow] ([xshift=-1.55cm]check.south) -- ++(0,-.38) -| (both.north);
\draw[arrow] ([xshift=1.55cm]check.south) -- ++(0,-.38) -| (one.north);
\draw[back] (both.south) |- (0,-3.15);
\draw[back] (one.south) |- (0,-3.15);
\fill[teal!70!black] (0,-3.15) circle (1.3pt);
\draw[back,->,>=stealth] (0,-3.15) -- (check.south);
\end{tikzpicture}
\caption{Revision of branching questions. Each branch must retain the target and at least one alternative. Only the target may satisfy both branches, and removing either visual clue must restore ambiguity. We repeat these checks after each revision and discard the question if no valid revision remains.}
\label{fig:branch_revision}
\end{figure*}

For the museum example in Figure~\ref{fig:query_construction}(b), we start with two collections of images: works by Jan van Eyck or his workshop, and works by Rembrandt. We ask the generator to inspect each collection independently and write a condition shared by several images. For each matching work, its source identifies the museum that holds it. We count each museum once, even when several of its works match.

\begin{systempromptbox}[Question construction prompt]
\raggedright\textbf{Initial branch description (excerpt)}\par
Construct a moderately selective visual condition that includes the target AND at least one genuine non-target candidate entity. Do not uniquely describe the target.
\end{systempromptbox}

The resulting conditions select a main figure wearing a floor-length red robe in the first collection and an oil painting in the second. Each condition retains works from several institutions.

\subsection{Composing Sequential and Branching Reasoning Steps}

\noindent\textbf{Sequential composition.}
We extract a directed relation from the source entity's Wikipedia article: the Red Bull RB8 competed against the Sauber C31. This relation connects the two visual steps in a three-step skeleton. The first photograph identifies the Sauber C31 at the Canadian Grand Prix, the text relation leads to a rival model, and the second photograph identifies the race in which that rival appears.

\begin{systempromptbox}[Question construction prompt]
\raggedright\textbf{Text relation (excerpt)}\par
Check the actual B-predicate-C statement against the evidence, including direction and roles. Do not collapse a multi-link explanation into an unsupported direct relation.
\end{systempromptbox}

\noindent\textbf{Branching composition.}
In the museum example, the first branch retains the Detroit Institute of Arts, the Louvre, and the National Gallery of Victoria. The second retains the Detroit Institute of Arts, the Metropolitan Museum of Art, and the Old Masters Picture Gallery in Dresden. Only Detroit occurs in both sets.

We check how much each visual condition contributes to identifying the answer. For branch $i$, we collect the entities supported by all inspected images in $U_i$ and those supported by matching images in $S_i$. Each $S_i$ must contain the target and at least one alternative, and $S_A\cap S_B$ must contain only the target. We then remove one visual condition at a time while keeping its source relation. In the museum example, $U_A\cap S_B$ also admits the Metropolitan Museum of Art, and $S_A\cap U_B$ also admits the Louvre. Both conditions therefore contribute to selecting Detroit within the inspected collections.

Figure~\ref{fig:branch_revision} shows how we use these checks to revise a question. When either condition can be removed without changing the answer, we ask the generator to write a new visual clue for each collection. We allow this joint rewrite at most once per question. When the intersection contains a competing entity, we can instead narrow one branch at a time, with up to four such attempts across both branches. Each accepted revision retains the target and at least one alternative from the previous set. We recompute the intersection after each revision and proceed to composition as soon as the checks pass.

\begin{systempromptbox}[Question construction prompt]
\raggedright\textbf{Narrowing one branch (excerpt)}\par
Lightly NARROW the previous condition using an actual visible qualifier to remove suggested conflicts. Preserve the target AND at least one other non-target entity.
\end{systempromptbox}

For example, one question combines portraits of leaders associated with the 48th G7 summit and photographs of leaders meeting Joe Biden. Initially both Emmanuel Macron and Joko Widodo matched both descriptions. On inspecting the portraits again, the generator added a dark necktie to the portrait condition. Widodo wore a red tie, so this detail removed him from the intersection. The original meeting description and the revised portrait description then jointly identified Macron.

\begin{figure*}[!t]
\centering
\includegraphics[width=\textwidth]{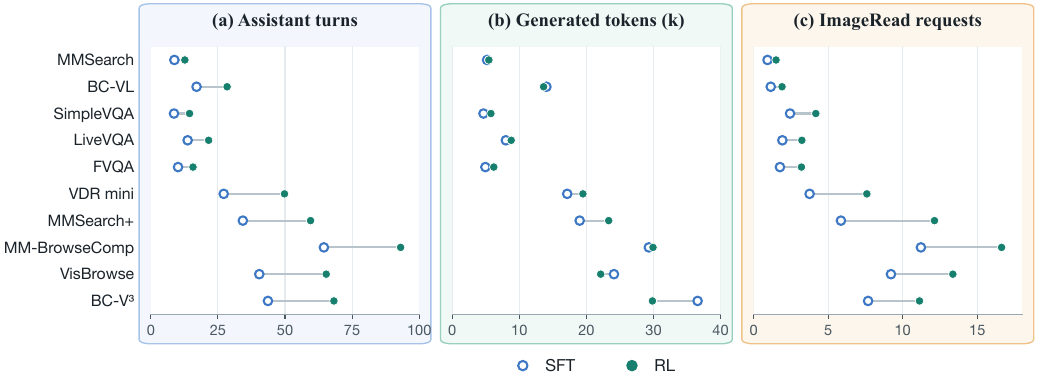}
\caption{Search behavior before and after RL on the same evaluation questions. Each point averages three runs within a benchmark. Assistant turns include the final response; generated tokens exclude tool observations. The right panel counts \texttt{ImageRead} requests, including unsuccessful requests.}
\label{fig:sft_rl_behavior}
\end{figure*}

\begin{figure*}[!t]
\centering
\includegraphics[width=\textwidth]{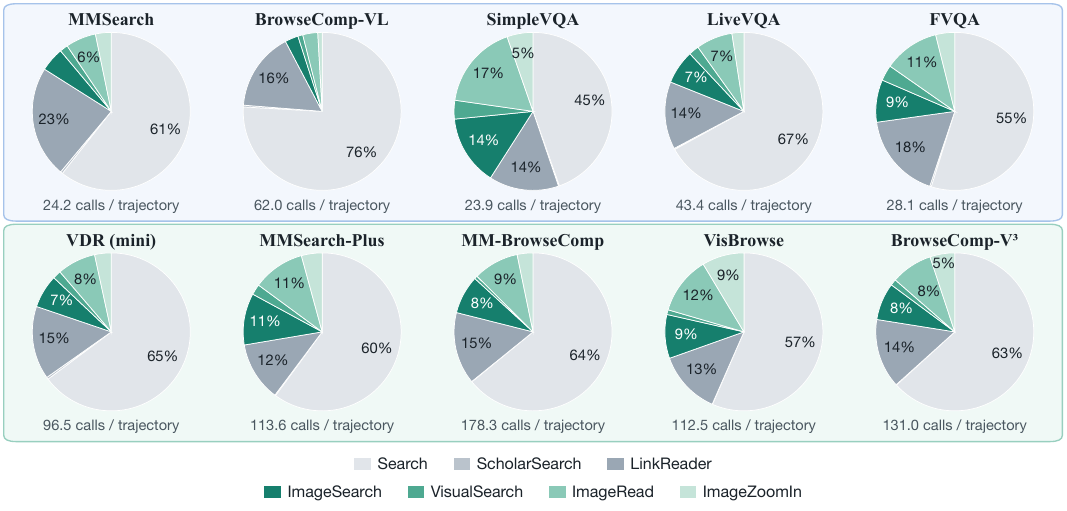}
\caption{Tool-use distributions of VGR-Searcher RL across ten benchmarks, pooled over three evaluation runs. Each pie shows the share of counted uses of the seven supported tools; the annotation below gives mean total tool use per trajectory. Percentage labels are omitted for shares below 5\%.}
\label{fig:rl_benchmark_tools}
\end{figure*}

\subsection{Writing and Verifying Complete Questions}

We link the public clues in each verified skeleton to form a question. For a sequential question, we name the starting entity and refer to each intermediate entity through the preceding clue. In the racing example, the phrase \emph{the aforementioned white and black race car} carries the result of the first visual step into the rivalry relation. The final clause asks for the race shown in the photograph of the rival car.

\begin{systempromptbox}[Question construction prompt]
\raggedright\textbf{Question composition (excerpt)}\par
For each T step express text\_relation faithfully, without adding a visual condition. Carry the OUTPUT ENTITY of each step into the next; an incidental object mentioned in a text relation is not automatically the next step's input.
\end{systempromptbox}

\begin{table*}[!t]
\centering\small
\caption{Interaction length of VGR-Searcher RL, pooled over three evaluation runs. Counts refer to trajectories, and turn means include the final answer. The correctness split excludes three unscorable MMSearch-Plus trajectories.}
\label{tab:rl_benchmark_length}
\begin{tabular}{@{}lrrrrrr@{}}
\toprule
& \multicolumn{2}{c}{All} & \multicolumn{2}{c}{Correct} & \multicolumn{2}{c}{Incorrect}\\
\cmidrule(lr){2-3}\cmidrule(lr){4-5}\cmidrule(l){6-7}
Benchmark & Count & Turns & Count & Turns & Count & Turns\\
\midrule
MMSearch & 900 & 12.7 & 795 & 10.4 & 105 & 29.7\\
BrowseComp-VL & 1,197 & 28.4 & 857 & 20.6 & 340 & 48.0\\
SimpleVQA & 900 & 14.4 & 751 & 9.4 & 149 & 39.9\\
LiveVQA & 900 & 21.5 & 814 & 17.0 & 86 & 64.9\\
FVQA & 900 & 15.7 & 783 & 13.0 & 117 & 34.0\\
\midrule
VDR (mini) & 1,500 & 49.8 & 681 & 34.2 & 819 & 62.8\\
MMSearch-Plus & 933 & 59.5 & 518 & 37.3 & 412 & 87.7\\
MM-BrowseComp & 672 & 93.1 & 324 & 72.3 & 348 & 112.4\\
VisBrowse & 507 & 65.4 & 341 & 53.4 & 166 & 89.8\\
BrowseComp-V$^3$ & 900 & 68.2 & 468 & 50.7 & 432 & 87.2\\
\bottomrule
\end{tabular}
\end{table*}

\noindent\textbf{Complete sequential question.}
\begin{quote}\small The 2012 Canadian Grand Prix featured a primarily white and black open-wheel race car driving on a paved race track, with a red circular logo on its white side panel, red trim on all its tires, and a solid black rear wing. A separate Formula One car that competed against the model of the aforementioned white and black race car appears as a single dark blue open-wheel racing car covered in fine yellow granular debris, with blue-accented tire rims, traveling along a track edge marked with white and red painted lines. Which Formula One Grand Prix event is this dark blue race car competing in?\par\textbf{Answer.} 2012 British Grand Prix.\end{quote}

For a branching question, we express the two conditions as parallel constraints on the same answer. The phrase \emph{holds both} asks for one museum, while each clause describes a different work in its collection.

\noindent\textbf{Complete branching question.}
\begin{quote}\small Which art museum institution holds both a painting created by Jan van Eyck or his workshop that features at least one main human figure wearing a floor-length full red robe, and an oil painting created by Rembrandt that is not a drawing, etching, or other printed medium?\par\textbf{Answer.} Detroit Institute of Arts.\end{quote}

Finally, we ask the construction model to check the wording of the complete question against the skeleton. It checks that each clause preserves its visual condition, refers to the intended entity, and supplies the input needed by the next step. We screen sequential questions for descriptions that disclose an intermediate entity or allow the final entity to be inferred from the relation alone. We also give the complete question to a solver without tools and filter questions it can answer directly. Table~\ref{tab:suppl_sft_data} compares SFT on the resulting sequential and branching questions.

\section{Search Behavior Across Benchmarks}
\label{app:search_behavior}
\subsection{Search Behavior Before and After RL}
We compare the SFT and RL policies on the same 3,103 questions across ten benchmarks, using three independent runs per policy (Figure~\ref{fig:sft_rl_behavior}). Generated tokens include reasoning, tool calls, and final answers, but exclude tool observations. Overall means weight the benchmarks equally.

After RL, mean interaction length increases by about 60\%, while total generation per question changes by only 1\% on average. Per-turn generation becomes shorter on all ten benchmarks, and image-reading requests become more frequent.

\subsection{Tool Use Across Multimodal Benchmarks}
We evaluate the RL policy at update 77 in three independent runs across all ten multimodal benchmarks. Each run contains 3,103 questions, yielding 9,309 trajectories in total. BrowseComp-VL combines its 199 L1 and 200 L2 questions, and VDR uses the 500-question mini split. Figure~\ref{fig:rl_benchmark_tools} pools the three runs and shows the distribution of tool use for each benchmark, together with the mean total per trajectory. A batched \texttt{Search} request contributes one count per query; other requests count once. We count successful and unsuccessful agent requests and exclude service-level retries.

Tool use varies substantially across benchmarks. MM-BrowseComp elicits the most extensive search among these evaluations, averaging 114.3 web queries and 16.6 \texttt{ImageRead} calls per trajectory. BrowseComp-V$^3$, MMSearch-Plus, and VisBrowse also average more than ten image inspections per trajectory, compared with 1.5 on MMSearch. Fine-grained inspection is especially frequent on VisBrowse, where \texttt{ImageZoomIn} averages 9.7 calls. At least one visual tool is used in 98.6\% of VisBrowse trajectories and 94.4\% of BrowseComp-V$^3$ trajectories, compared with 40.1\% on MMSearch. These rates measure tool use beyond the images already present in the question.

\subsection{Interaction Length and Answer Correctness}
Table~\ref{tab:rl_benchmark_length} reports assistant turns, including the final response, for all trajectories and separately for correct and incorrect answers. Final-answer judgments determine group membership.

Incorrect answers are associated with longer interactions on every benchmark. On BrowseComp-V$^3$, correct trajectories average 50.7 turns and incorrect trajectories average 87.2; on MMSearch-Plus, the corresponding values are 37.3 and 87.7. Even on MMSearch, where the overall mean is only 12.7 turns, incorrect trajectories average 29.7 turns.


\section{Qualitative Search Trajectories}
\label{app:qualitative_search}

\begin{figure*}[p]
\centering
\includegraphics[width=\textwidth]{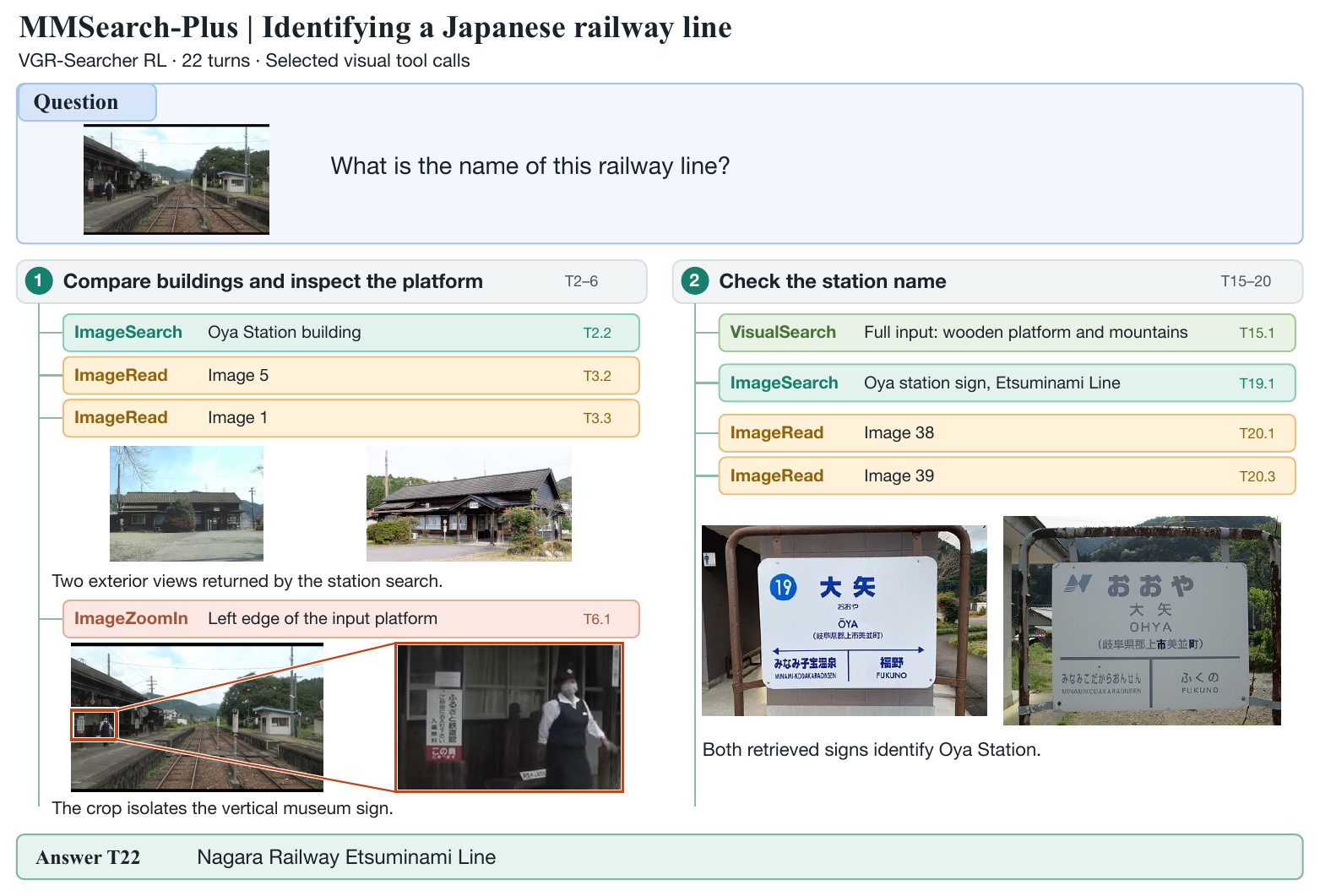}
\caption{Identifying a railway line from a platform photograph. The orange guides connect the selected region to the crop returned at turn 6. Turn numbers mark selected steps in the recorded trajectory.}
\label{fig:trace_railway}
\end{figure*}

\begin{figure*}[p]
\centering
\includegraphics[width=\textwidth]{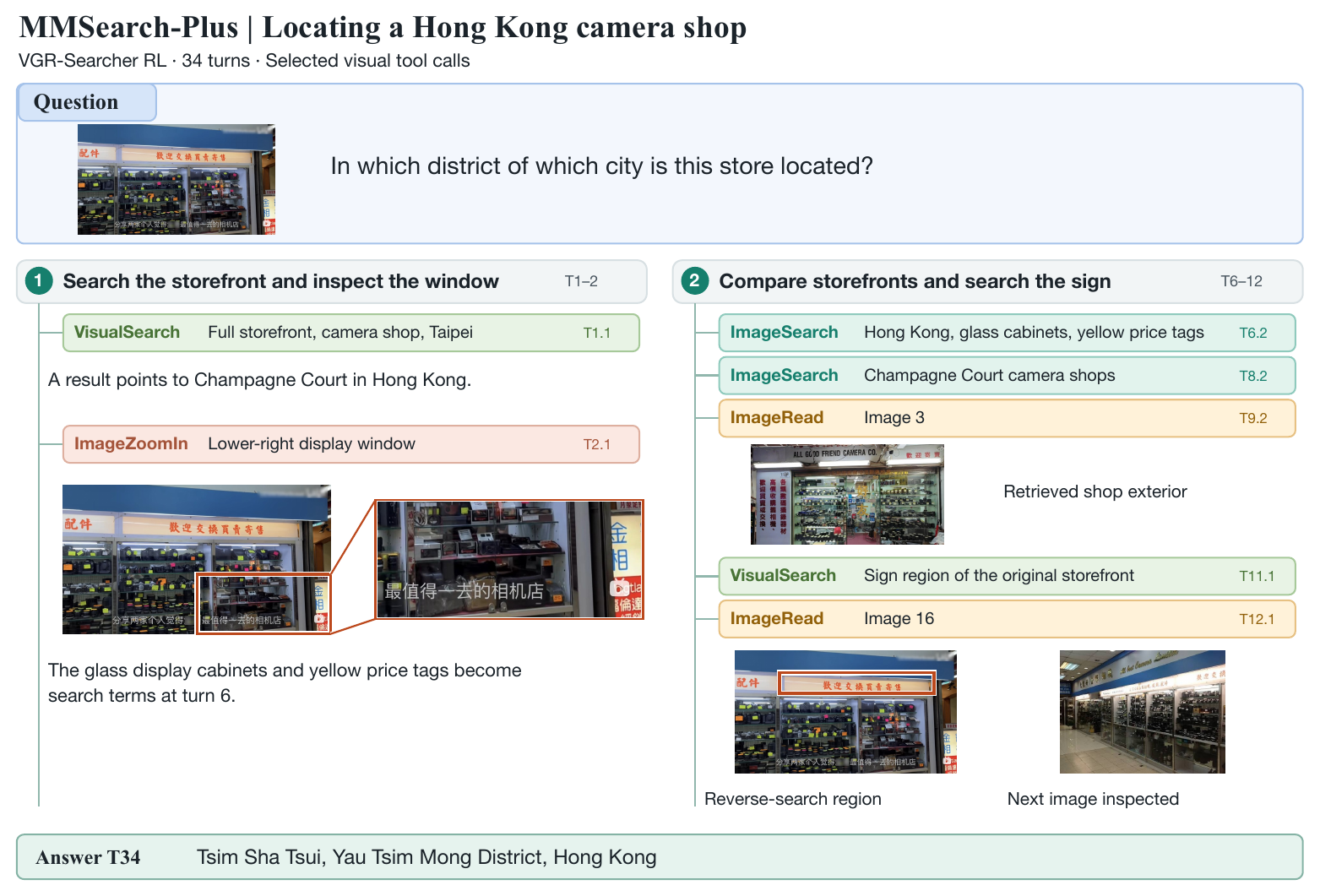}
\caption{Locating a camera shop from its storefront. The linked panels at turn 2 show the crop and its source. The box at turn 11 marks the region submitted for reverse image search.}
\label{fig:trace_camera}
\end{figure*}

We illustrate how VGR-Searcher uses visual evidence in five correctly answered questions across four benchmarks. In one MMSearch-Plus example, the agent inspects a platform sign and retrieves station photographs to identify the railway line (Figure~\ref{fig:trace_railway}). In the other, details in a shop window guide its search for a camera store in Hong Kong (Figure~\ref{fig:trace_camera}). On VisBrowse, it follows collection records to locate a painting and compares details across reproductions (Figure~\ref{fig:showcase_long}). On BrowseComp-V$^3$, it traces an artwork through publication records and inspects a related work (Figure~\ref{fig:showcase_art}). On MM-BrowseComp, it follows textual clues to a game guide and compares object shapes in the guide images (Figure~\ref{fig:showcase_rocks}).

\begin{figure*}[p]
\centering
\includegraphics[width=\textwidth]{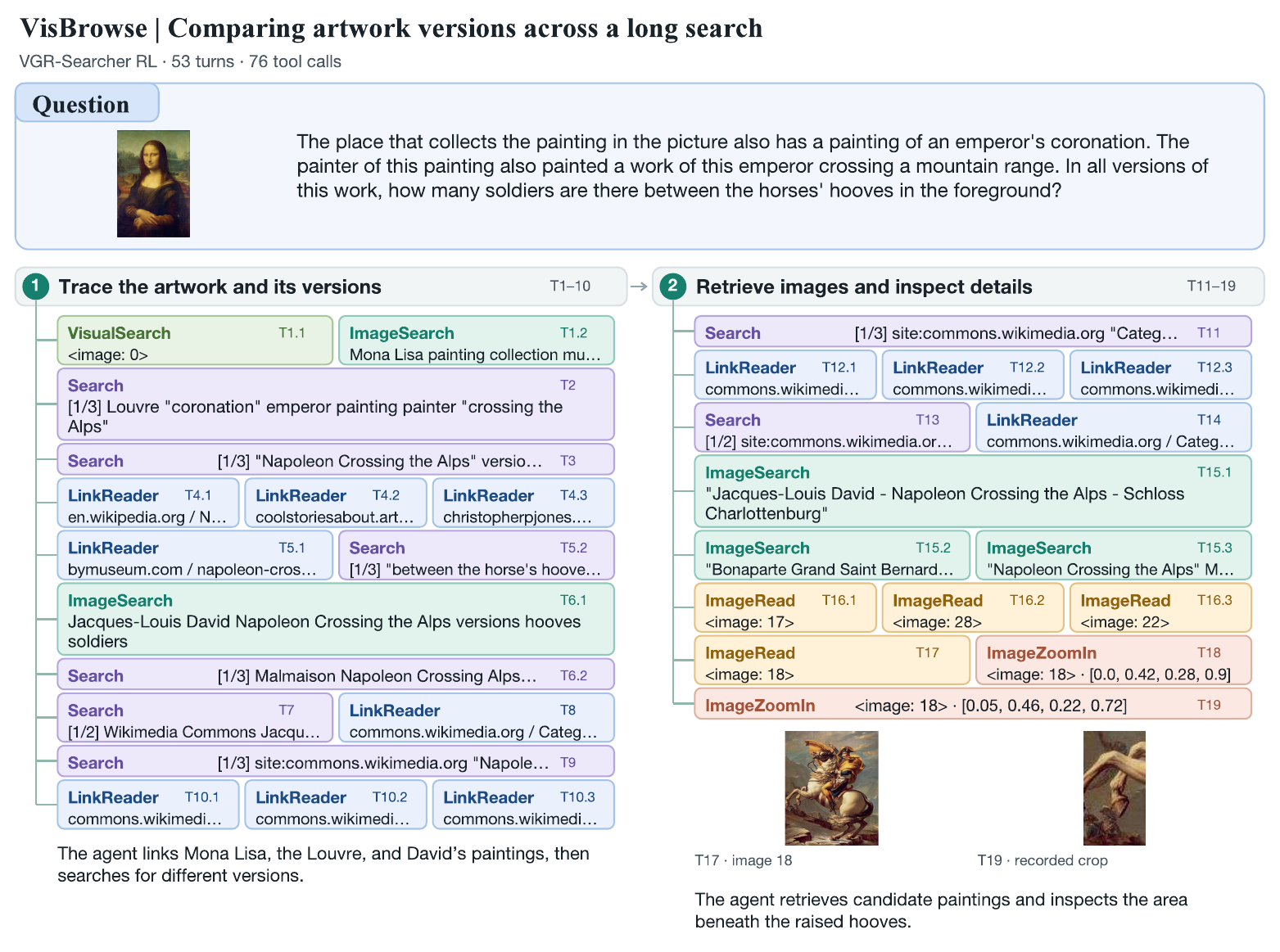}
\caption{Complete VisBrowse trajectory with 53 assistant turns and 76 tool calls (part 1 of 2). The agent follows collection and artist relations, retrieves candidate images, and begins inspecting local details. Stages and calls are read left to right, then top to bottom.}
\label{fig:showcase_long}
\end{figure*}

\begin{figure*}[p]
\ContinuedFloat
\centering
\includegraphics[width=\textwidth]{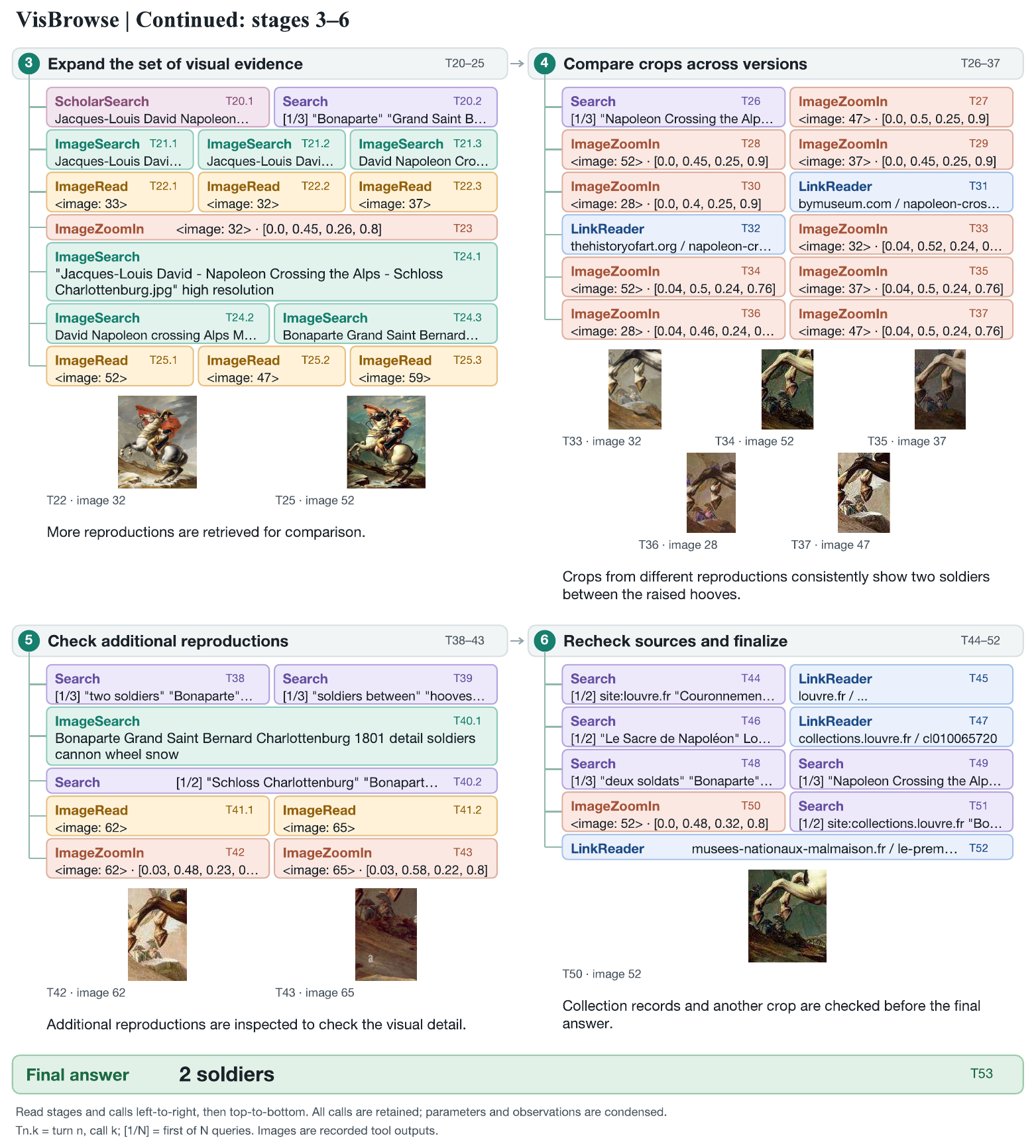}
\caption{VisBrowse trajectory (continued). The agent expands its image evidence, compares crops across reproductions, and revisits source records before answering. Images: benchmark input, Wikimedia Commons, and Alamy.}
\end{figure*}

\begin{figure*}[p]
\centering
\includegraphics[width=\textwidth]{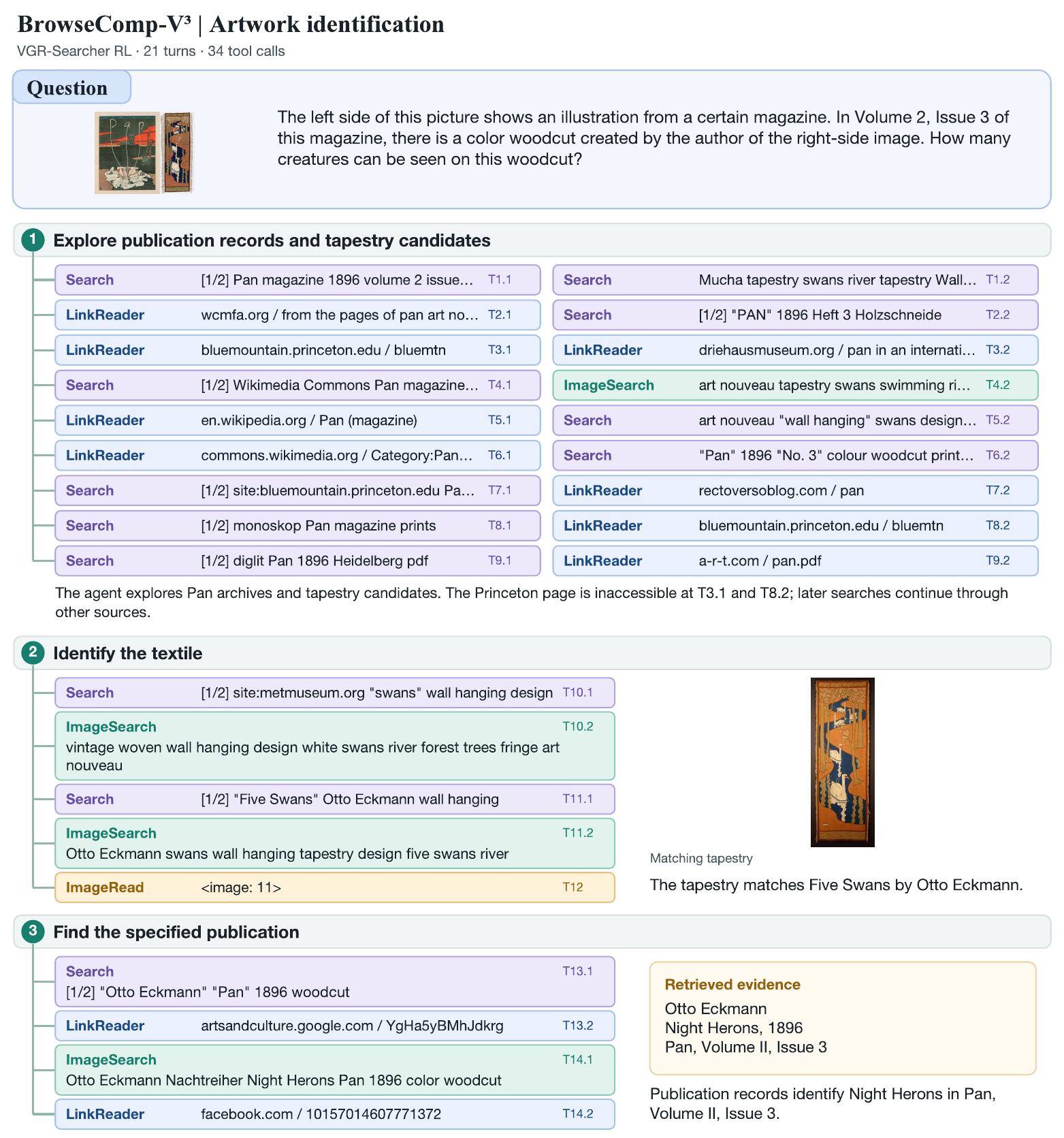}
\caption{Complete tool-call sequence on BrowseComp-V$^3$ (part 1 of 2). The agent explores publication records, identifies an artwork, and retrieves a related work. Each coloured box is one tool call; T$n.k$ denotes call $k$ in assistant turn $n$. The boxes show abbreviated arguments.}
\label{fig:showcase_art}
\label{fig:visual_tools}
\end{figure*}

\begin{figure*}[p]
\ContinuedFloat
\centering
\includegraphics[width=\textwidth]{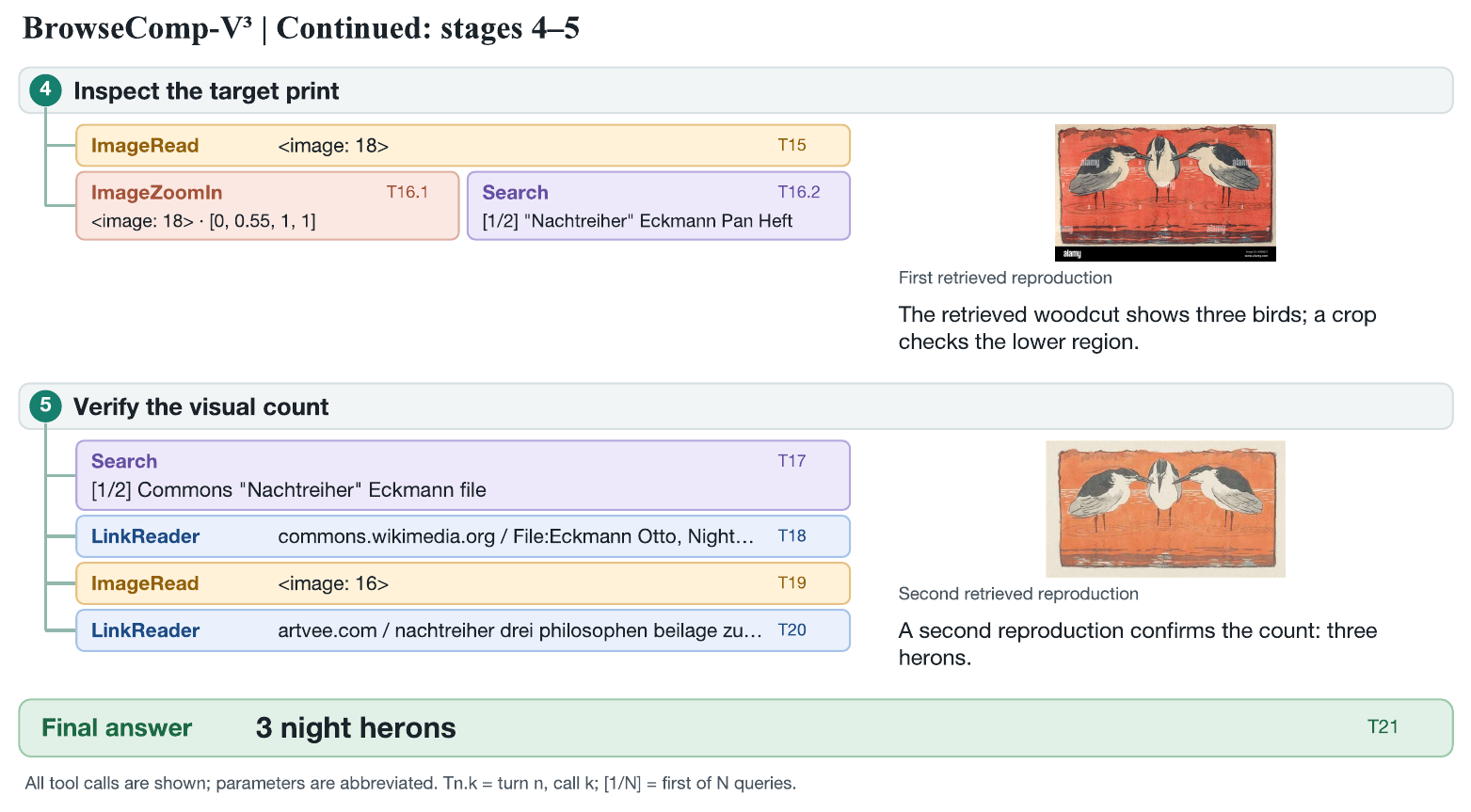}
\caption{BrowseComp-V$^3$ trajectory (continued). The agent inspects the retrieved work and checks another reproduction before answering. Images: benchmark input, Wikimedia Commons / Br\"ohan Museum, Alamy, and MutualArt.}
\end{figure*}

\begin{figure*}[p]
\centering
\includegraphics[width=\textwidth]{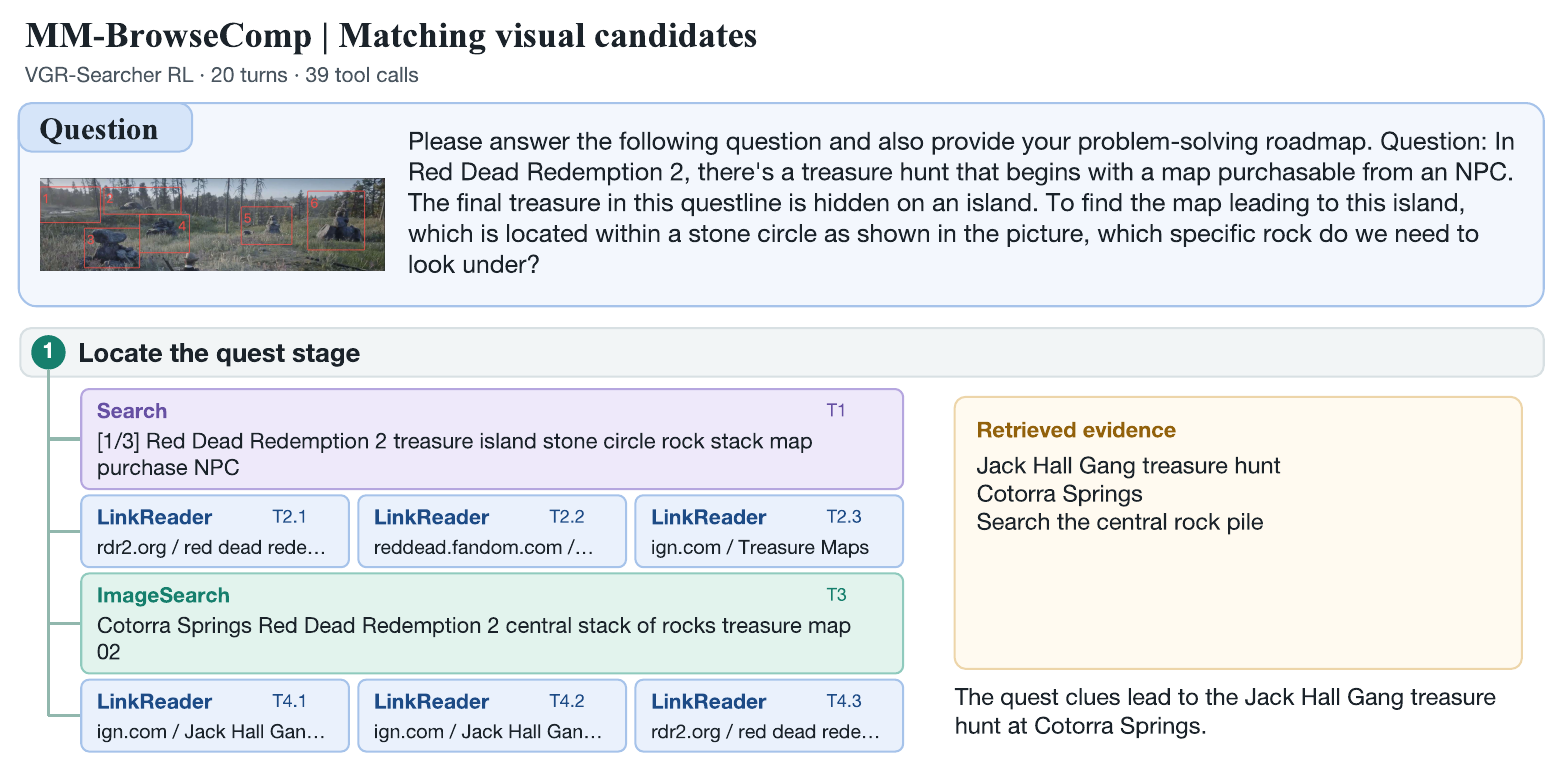}
\caption{Complete tool-call sequence on MM-BrowseComp (part 1 of 2). The agent uses the question's textual clues to locate the relevant game quest and collect guide material.}
\label{fig:showcase_rocks}
\end{figure*}

\begin{figure*}[p]
\ContinuedFloat
\centering
\includegraphics[width=\textwidth]{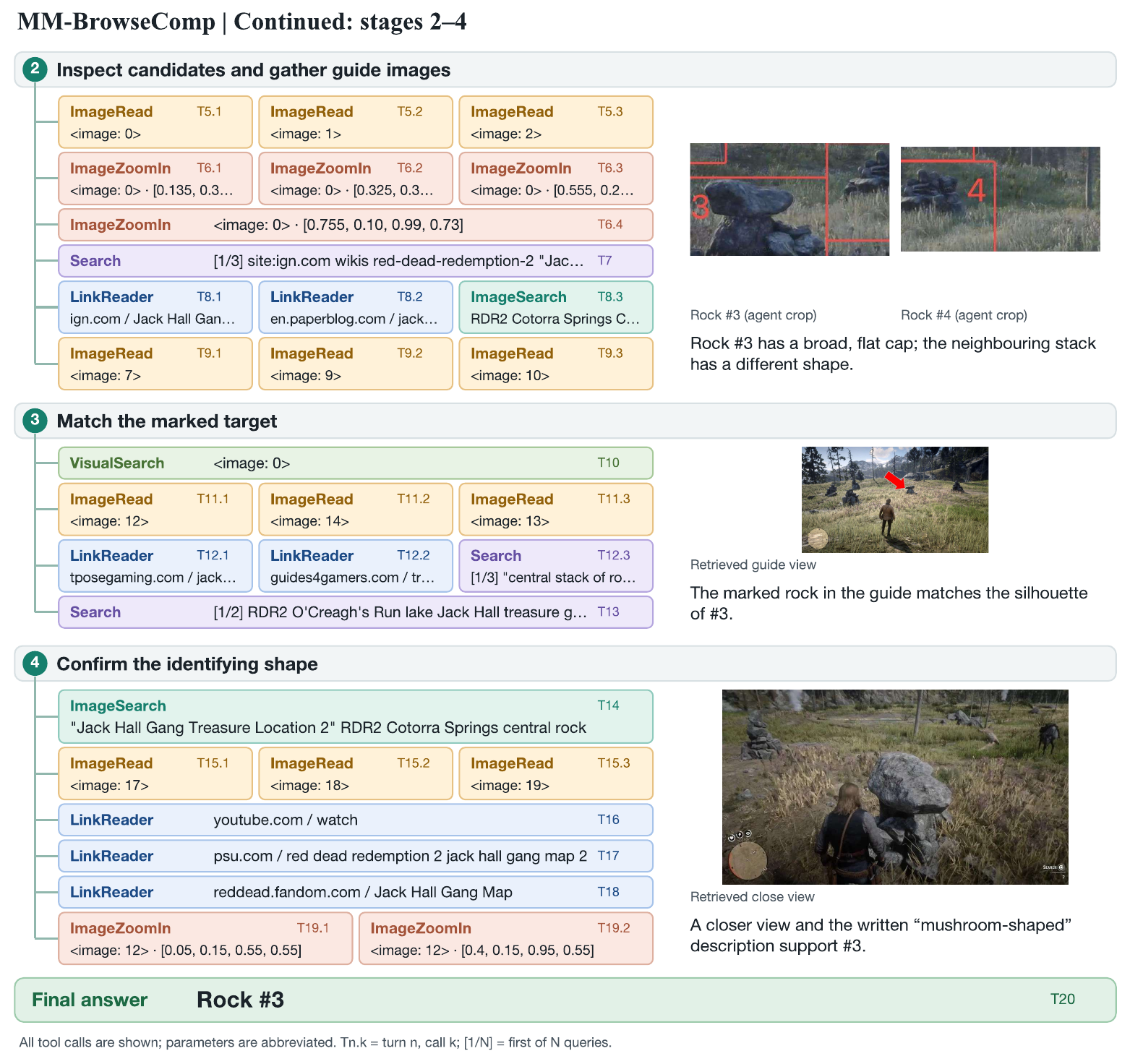}
\caption{MM-BrowseComp trajectory (continued). The agent compares candidate objects with guide images, using crops and further retrieval to check distinguishing shapes. Game imagery: Rockstar Games; retrieved guides: Guides4Gamers and IGN.}
\end{figure*}

\clearpage

\section{Code and Data}
\label{app:code_data}

The accompanying files contain the following code and example questions.
\begin{itemize}
    \item \texttt{code\_implementation/evaluation/}: the evaluation framework for multimodal search agents.
    \item \texttt{code\_implementation/construct\_MMSAQA/}: the multimodal deep search question construction pipeline.
    \item \texttt{data\_samples/}: two JSON files containing 20 sequentially composed questions and 20 branching questions, respectively.
\end{itemize}